\documentclass[letterpaper]{article} 
\usepackage{aaai2027}  
\usepackage[hyphens]{url}  
\usepackage{graphicx} 
\usepackage{natbib}  
\usepackage{caption} 
\usepackage{algorithm}
\usepackage{algorithmic}

\usepackage{newfloat}
\usepackage{listings}
\DeclareCaptionStyle{ruled}{labelfont=normalfont,labelsep=colon,strut=off} 
\floatstyle{ruled}
\newfloat{listing}{tb}{lst}{}
\floatname{listing}{Listing}

\usepackage{booktabs}

\usepackage{algorithm}
\usepackage{algorithmic}
\usepackage{multirow}
\usepackage{comment}
\usepackage{booktabs}
\usepackage{multirow}
\usepackage{makecell}
\usepackage{tikz}
\usepackage{subcaption}
\usepackage{amsmath}
\usepackage[export]{adjustbox}
\usepackage{xurl}

\title{Toward a More Ethical Facial Age Estimation: \\ A Generalized Zero-Shot Benchmark Without Training on Children's Data}
\author{
    Caio Petrucci\textsuperscript{\rm 1}, Leo Sampaio Ferraz Ribeiro\textsuperscript{\rm 2}, 
    Sandra Avila\textsuperscript{\rm 1}
}
\affiliations{
    \textsuperscript{\rm 1}Instituto de Computação, Universidade Estadual de Campinas (UNICAMP), Campinas, Brazil\\
    \textsuperscript{\rm 2}Instituto de Ciências Matemáticas e de Computação, Universidade de São Paulo (USP), São Carlos, Brazil\\
    caio.rosa@students.ic.unicamp.br, leo.ribeiro@icmc.usp.br, sandra@ic.unicamp.br
}

\begin{document}

\maketitle

\begin{abstract}
Age estimation from facial images typically relies on training data that includes images of minors, a practice that raises ethical, legal, and privacy concerns and that child-data governance frameworks explicitly advise against. While the task remains relevant (e.g., for detecting child sexual abuse imagery), we advocate against using data from minors entirely and quantify what the exclusion costs in accuracy. We formalize age estimation without children's training data as a \textit{generalized zero-shot learning} (GZSL) problem: age intervals present during training are \textit{seen} classes and withheld intervals are \textit{unseen}, with models evaluated jointly on both. The generalized setting, rather than conventional zero-shot evaluation on unseen classes alone, is the appropriate one here because a deployed estimator must operate across the entire lifespan, not only on the interval withheld from it. Revisiting six widely used datasets, we introduce standardized splits with strict age-group separation. For datasets with identity annotations, subject-age-exclusive splits prevent identity leakage across the seen/unseen boundary. Evaluating nine state-of-the-art age estimation methods under this protocol reveals that all of them fail to generalize to unseen age groups, suffering substantial degradation --- on average 46.4\%, and up to 52.8\% --- relative to the supervised baseline. Moreover, models do not simply degrade: they systematically anchor predictions for unseen ages to nearby seen classes, a manifestation of the well-known seen-class bias in generalized zero-shot learning.
\end{abstract}

\begin{links}
   \link{Code \& datasets}{https://github.com/caiopetruccirosa/generalized-zero-shot-age-estimation}
\end{links}

\section{Introduction}

Facial age estimation is among the most widely adopted computer vision tasks, and its literature has produced numerous public training and evaluation datasets~\cite{ricanek2006morph, moschoglou2017agedb, chunchen2014cacd, agustsson2017apparentrealage, zhang2017utkface}. These datasets differ substantially in collection procedures, annotation protocols, demographic composition, and image characteristics --- ranging from highly controlled mug-shot imagery to unconstrained \textit{in-the-wild} scenarios with varying pose, illumination, expression, and quality. Compounding this, evaluation protocols remain poorly standardized, with split design and subject overlap having a substantial impact on reported performance~\cite{paplhamcvpr2024reflect}.

One application gives the task unusual stakes. The volume of child sexual abuse imagery (CSAI) circulating online overwhelms the agents responsible for triaging it, and automating that triage cannot proceed by ordinary means: the target data is legally restricted, and minimizing its contact with researchers and models is a governing constraint. \citet{coelho2025proxytasks} formalize the resulting pattern as \textit{Proxy Tasks}; Sub-tasks explored under this framing include scene understanding, nudity detection, and age estimation; the last two have absorbed most of the effort, since CSAI is the conjunction of sexual content with a minor subject and nudity detection alone cannot separate it from legal adult material.

That literature has adopted the sub-task without the restriction. Confronted with poor accuracy on minors, it responded by collecting more children's faces~\cite{anda2020deepuage, castrillon2018nonadult} --- which \citet{coelho2025proxytasks} note conflicts with the Responsible Data for Children framework~\cite{young2019rd4c}, though they set the question aside as outside their scope. We take it up. Adopting the recommendation makes the task harder, and that is the point: the cost of following it has never been measured, so the field has no basis on which to weigh the guideline against the accuracy it forfeits.

\begin{figure*}
    \centering
    \includegraphics[width=\linewidth]{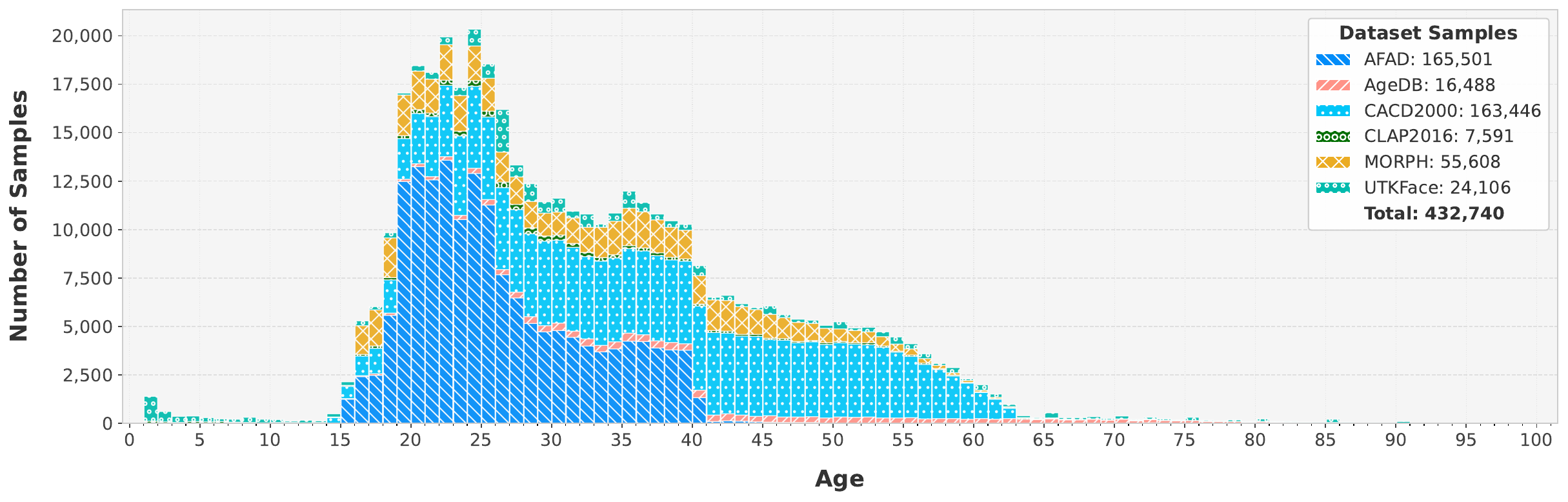}
    \caption{Age distribution across facial age estimation datasets --- AFAD, AgeDB, CACD2000, CLAP2016, UTKFace, and MORPH --- that compose our benchmark. Although the distributions vary substantially across datasets due to differences in collection protocols and demographic composition, children and adolescent age ranges remain consistently represented across widely adopted facial age estimation benchmarks.}
    \label{fig:age-distribution-histogram}
\end{figure*}

The concern is not confined to the forensic setting. Web-scraped corpora underpinning modern computer vision repeatedly contain children's images gathered without consent: Human Rights Watch identified Brazilian children in LAION-5B~\cite{schuhmann2022laion5b, hrw2024brazilchildrenlaion} with metadata exposing their home addresses, and \citet{caetano2025neglectedrisks} found children in 15.39\% of \#PraCegoVer \cite{pracegover}, an Instagram-scraped captioning dataset. \citet{birhane2021multimodal} and later The Stanford Internet Observatory were able to identify suspected CSAI within LAION-5B itself, prompting a filtered re-release~\mbox{\cite{thiel2023csam, laion2024relaion}}. 

Such failures do not stay at collection time. These corpora propagate through pretrained components --- LAION-5B underpins Stable Diffusion and OpenCLIP~\cite{rombach2022diffusion, cherti2023openclip}, which in turn seed much of the vision-language stack --- and what the data contains resurfaces as capability: in January 2026, Ofcom opened a formal investigation into X after Grok was used to generate undressed images of individuals, including sexualized synthetic images of minors~\cite{ofcom2026grok}. Remediation lags behind, since a filtered re-release does not retrain the models already built on the original. The point of control is the training set. Age estimation datasets are no exception: children appear throughout them (Figure~\ref{fig:age-distribution-histogram}) and are used in training without restriction.

We formulate age estimation under this restriction as a \textit{Generalized Zero-Shot Learning} (GZSL) problem: age intervals available in training are \textit{seen}, those excluded \textit{unseen}. We adopt GZSL rather than conventional ZSL --- which evaluates only unseen classes --- because deployed estimators must operate across the whole lifespan~\cite{xian2018zslthegoodbadugly}. Adults aged 18--59 are the seen classes; elderly individuals 60+ are unseen validation classes; children under 18 are unseen \mbox{test classes.} 

We propose a benchmark on six widely used datasets --- AFAD, AgeDB, CACD2000, CLAP2016, UTKFace, and MORPH --- defining standardized splits that jointly enforce age and subject exclusivity, so that neither identity leakage nor seen/unseen age overlap can inflate results. We then evaluate nine state-of-the-art methods spanning classification, Label Distribution Learning, and rank learning. All nine degrade sharply on unseen ages while remaining competitive on seen ones, and the degradation is not random: predictions for unseen ages are systematically pulled toward the nearest seen class. Across every family we tested, current age estimation paradigms rely on supervised signal within the observed age range and do not extrapolate beyond it.

Our contributions are: (i)~a formalization of age estimation without children's training data as a GZSL benchmark, supplying the restriction component that the proxy-task literature has so far omitted; (ii)~standardized subject-age-exclusive splits across six datasets, together with the algorithm that constructs them; (iii)~baselines for nine methods under this protocol, establishing where the current literature stands; and (iv)~an analysis of the resulting failure mode, which we identify as the seen-class bias familiar from GZSL.

\section{Related Work}
\subsection{Facial Age Estimation}
\label{sec:work-facial-age-estimation}

Facial age estimation methods broadly fall into three categories: (i)~classification and regression, (ii)~Label Distribution Learning, and (iii)~ordinal regression and rank learning.

Early deep learning approaches formulated age estimation as either regression or classification. Regression methods predict a continuous age directly through losses such as MSE or MAE; classification methods discretize ages into categories and optimize cross-entropy. DEX~\cite{rothe2015dex} exemplifies the latter, computing the final prediction as the expected value over predicted class probabilities. Classification often performs strongly but ignores the ordinal nature of age labels, treating neighboring ages as independent categories.

To incorporate age continuity, Label Distribution Learning (LDL) methods~\cite{gao2017dldl, gao2018dldlv2, diaz2019softlabels, pan2018meanvarianceloss} represent annotations as distributions centered on the target age, so neighboring labels receive non-zero probability. Ordinal regression and rank learning instead model relative ordering explicitly: OR-CNN~\cite{niu2016multiplecnn} casts estimation as binary ranking problems over predefined thresholds, while CORAL~\cite{cao2020coral} and CORN~\cite{shi2023corn} add consistent ranking formulations for stability.

Despite this progress, all three families remain fundamentally supervised, assuming every relevant age interval is represented during training --- an assumption that fails exactly when a specific age group cannot be used during \mbox{model development.}

Several public datasets support the task, differing in collection protocol, demographics, image conditions, and annotation methodology (Table~\ref{tab:datasets}). Two properties matter for what follows. First, all six contain children (Figure~\ref{fig:age-distribution-histogram}). Second, their age labels are largely \textit{not} ground truth: AFAD derives labels from the difference between a user's stated date of birth and photo upload date, CACD2000 from celebrity birth years paired with search-engine image timestamps, and CLAP2016 supplements real age with crowd-sourced \textit{apparent} age. Only AgeDB is manually annotated throughout, and MORPH's labels come from institutional records. Reported errors on these datasets therefore include a dataset-dependent component of label noise, a point we return to when interpreting results.

Overall, these datasets differ substantially not only in demographic composition and image conditions but also in collection procedures and annotation reliability. For instance, AFAD predominantly represents East Asian individuals, whereas CACD2000 and AgeDB are strongly influenced by celebrity imagery. Similarly, MORPH presents a particularly distinct social context due to its mug-shot acquisition protocol. Such differences reinforce the importance of standardized and carefully controlled evaluation protocols when comparing methods across datasets.

\begin{table}[t]
    \centering
    \footnotesize
    \renewcommand{\arraystretch}{1.25}
    \setlength{\tabcolsep}{3pt}
    \begin{tabular}{lrrcl}
        \toprule
        & \textbf{Images} & \textbf{Subj.} & \textbf{Ages} & \textbf{Age labels} \\
        \midrule
        \textbf{AFAD}     & 165,501 & 149,918 & 15--40 & Inferred \\
        \textbf{AgeDB}    & 16,488  & 568    & 1--101  & Verified \\
        \textbf{CACD2000} & 163,446 & 2,000  & 16--62  & Inferred \\
        \textbf{CLAP2016} & 7,591   & ${\sim}7{,}000^{*}$ & 0--95 & \makecell[l]{Real + \\ Crowdsourced} \\
        \textbf{UTKFace}  & 24,106  & ---    & 0--116  & Estimated \\
        \textbf{MORPH}    & 55,608  & 13,617 & 16--77  & Real \\
        \bottomrule
    \end{tabular}
    \caption{Datasets composing the benchmark and how their age labels were obtained, which bounds how precisely error can be measured on each. \textbf{Real}: actual age, from custodial records (MORPH) or dataset metadata (CLAP2016). \textbf{Verified}: actual age, manually confirmed against known birth and photograph dates. \textbf{Inferred}: actual age approximated from a self-declared or published birth date and an upload or search year; CACD2000's only to the year. \textbf{Estimated}: inherited or web-scraped, never checked. \textbf{Crowdsourced}: apparent age judged from appearance over ${\sim}300{,}000$ votes, not a real age. $^{*}$CLAP2016 reports ${\sim}7{,}000$ individuals but no per-image identity, so it cannot be partitioned by subject.}
    \label{tab:datasets}
\end{table}

\subsection{Zero-Shot Learning}

Zero-Shot Learning (ZSL) recognizes classes never observed in training~\cite{xian2018zslthegoodbadugly}, transferring knowledge from seen to unseen classes via auxiliary information. Research centers on classification benchmarks --- AWA2, CUB~\cite{wah2011cub}, SUN~\cite{patterson2012sun} --- supplying predefined splits with per-category semantic side information, so methods target discrete categories: visual--semantic compatibility functions~\cite{frome2013devise, akata2013ale, akata2015sje}, attribute descriptors~\cite{lampert2014dapiap}, and generative feature synthesis~\cite{xian2018fclswgan, schonfeld2019cadavae}.

Two differences separate our setting. First, ZSL developed around classes that are rare or novel on use --- data that is hard to obtain; here the data exists and is abundant, and we decline to use it. Second, conventional ZSL targets classification, whereas age estimation is ordinal regression, so standard benchmark assumptions do not transfer. That is also an opportunity: age labels carry intrinsic ordinal structure, so the auxiliary information ZSL usually imports from external attributes is already present in the label space. What is missing is the protocol: the field still evaluates almost exclusively under supervised splits where train and test share an age distribution, so there is no standardized way to measure generalization to a withheld age interval, and no common ground on which methods designed to transfer across one could be compared.

\section{Generalized Zero-Shot Learning Benchmark}
\label{subsec:gzsl-benchmark}

We propose a novel benchmark for facial age estimation under the \textit{Generalized Zero-Shot Learning} (GZSL) setting, specifically designed to develop and evaluate age estimation methods without usage of children's images during training. Unlike conventional protocols, where training and testing samples share the same age distribution, we consider a setting in which images of children are completely removed from the training process, and put into the test set exclusively to evaluate generalization capability. Our benchmark 
treats age estimation as a \textit{zero-shot learning} problem with respect to children's data, as \textit{unseen} classes.

We adopt the \textit{generalized} ZSL formulation rather than conventional ZSL, which evaluates only unseen classes at test time. As discussed by \citet{xian2018zslthegoodbadugly}, GZSL better reflects real-world scenarios, since models must simultaneously operate on both seen and unseen classes during inference. By treating age estimation as a ZSL task to a specific age interval, in this case, ages below 18, it is crucial to model the problem considering the GZSL formulation. In our benchmark, seen adult age classes remain present during validation and testing, evaluated jointly with unseen age classes  in the same setting.

The benchmark enforces \textit{subject-exclusivity}: each individual is restricted to a single split. Inconsistently adopted in the literature, this is essential to prevent leakage~\cite{paplhamcvpr2024reflect} and ensure models learn aging patterns rather than identity-specific features. Under GZSL we extend the constraint to the seen/unseen age threshold within each evaluation split, requiring that no subject contribute images to both components simultaneously. The reason is the joint evaluation GZSL entails: if a subject's adult images sit in the seen evaluation set while their child or elder images sit in the unseen set, the model gains an implicit identity anchor unavailable in a principled zero-shot scenario. We call these two levels jointly \textit{subject-age exclusivity}.

A valid ZSL formulation requires strict seen/unseen separation: unseen ages must be absent from the validation data used for model selection as well as from training, since exposing them leaks information into development~\cite{xian2018zslthegoodbadugly}. Validating on seen classes alone, however, says nothing about zero-shot behavior. We therefore reserve a second age range as \textit{unseen} validation classes, giving three \mbox{disjoint groups:}

\begin{itemize}
    \item \textit{Seen} classes: adult individuals between 18 and 59 years old, present in the training, validation, and testing sets;
    \item Validation \textit{unseen} classes: elderly individuals older than 60 years;
    \item Test \textit{unseen} classes: children and adolescents younger than 18 years old.
\end{itemize}

Our choice of the elderly as validation \textit{unseen} classes follows the constraint that a valid ZSL protocol requires unseen classes to be absent from validation as well as training~\cite{xian2018zslthegoodbadugly}; children are disallowed since exposing them during model selection would leak the information the benchmark withholds. The elderly are then the only age interval remaining outside the adult training range. The choice is also structurally apt: like children, the elderly sit at a boundary of the age spectrum, so predicting them requires extrapolating aging patterns rather than interpolating between observed ages. Both groups pose the same \textit{kind} of problem.

We organize each dataset into five \textsc{folders}: \textsc{folder}~0 (seen training), 1 and 2 (seen validation and test), 3 and 4 (unseen validation and test). Without per-image identity annotations a straightforward age-based assignment suffices; this is the case for UTKFace, and for CLAP2016, which is reported to contain roughly 7,000 individuals but supplies no identity label per sample, leaving no basis on which to group images by subject. With them the assignment is non-trivial, since one individual may span several age groups and hence several \textsc{folders}; for these datasets we use a greedy algorithm satisfying \textit{subject-age exclusivity}, summarized below and given in full as Algorithm~1 of the \mbox{supplementary material}.

Each \textsc{folder} gets a \textit{target count} $t_k$: the images it would hold absent any exclusivity constraint. Subjects spanning a single age group are assigned unambiguously; for \textit{mixed} subjects we enumerate candidate (split, kept age group) pairs and take the one maximizing $n_c \cdot \max(0,\, t_c - r_c)/t_c^2$, where $n_c$ is images retained and $r_c$ the \textsc{folder}'s running count. The numerator measures remaining demand, and dividing by $t_c^2$ normalizes urgency by \textsc{folder} size so that small unseen \textsc{folders} outrank the far larger training one. Images outside the kept group are discarded. A correction sweep then fills any under-target unseen \textsc{folder}, and adult-only subjects are split $80\%/10\%/10\%$.

How much data this costs varies sharply with dataset structure (Table~\ref{tab:gzsl-benchmark-splits}). AFAD, CLAP2016, and UTKFace lose essentially nothing: the latter two carry no usable identity annotations, and AFAD averages 1.1 images per profile, so almost no subject can straddle an age boundary. CACD2000 discards 12.6\%, since its 2,000 celebrities average 81.7 images each across wide age spans. MORPH loses only 2.9\% despite being longitudinal, because its child portion covers just ages 16--17. AgeDB is the extreme case at 65.6\%: 568 subjects spanning ages 1--101 means many individuals straddle the seen/unseen boundary, and 70.83\% of what survives falls in unseen validation, leaving just 1,270 training images. Demographic composition constrains the splits independently --- AFAD and MORPH contribute only 154 and 338 unseen-validation samples respectively, both having sparse coverage above age 60.

\begin{table}[t]
    \centering
    \scriptsize
    \renewcommand{\arraystretch}{1.35}
    \setlength{\tabcolsep}{2.25pt}
    \begin{tabular}{lrrrrrrr}
        \toprule
        & \multirow{2}{*}{\textbf{Train}} & \multicolumn{2}{c}{\textbf{Validation}} & \multicolumn{2}{c}{\textbf{Test}} & \multirow{2}{*}{\textbf{Total}} & \multirow{2}{*}{\textbf{Disc.}} \\
        \cmidrule(lr{0pt}){3-4} \cmidrule(lr{0pt}){5-6}
        & & \textbf{Seen} & \textbf{Unseen} & \textbf{Seen} & \textbf{Unseen} & & \\
        \midrule
        \textbf{AFAD}\rule{0pt}{4ex}\rule[-2.5ex]{0pt}{0pt}     & \makecell{127,315 \\ \scriptsize{\textit{76.95\%}}} & \makecell{15,920 \\ \scriptsize{\textit{9.62\%}}}  & \makecell{154 \\ \scriptsize{\textit{0.09\%}}}    & \makecell{15,912 \\ \scriptsize{\textit{9.62\%}}}  & \makecell{6,154 \\ \scriptsize{\textit{3.72\%}}}  & 165,455 & 46 \\
        \textbf{AgeDB}\rule{0pt}{4ex}\rule[-2.5ex]{0pt}{0pt}    & \makecell{1,270 \\ \scriptsize{(\textit{22.39\%})}}   & \makecell{166 \\ \scriptsize{\textit{2.93\%}}}    & \makecell{4,018 \\ \scriptsize{\textit{70.83\%}}} & \makecell{124 \\ \scriptsize{\textit{2.19\%}}}     & \makecell{95 \\ \scriptsize{\textit{1.67\%}}}     & 5,673   & 10,815 \\
        \textbf{CACD2000}\rule{0pt}{4ex}\rule[-2.5ex]{0pt}{0pt} & \makecell{109,452 \\ \scriptsize{(\textit{76.59\%})}} & \makecell{13,732 \\ \scriptsize{\textit{9.61\%}}} & \makecell{2,912 \\ \scriptsize{\textit{2.04\%}}}  & \makecell{13,561 \\ \scriptsize{\textit{9.49\%}}} & \makecell{3,242 \\ \scriptsize{\textit{2.27\%}}}  & 142,899 & 20,547 \\
        \textbf{CLAP2016}\rule{0pt}{4ex}\rule[-2.5ex]{0pt}{0pt} & \makecell{4,839 \\ \scriptsize{(\textit{63.76\%})}}   & \makecell{605 \\ \scriptsize{\textit{7.97\%}}}    & \makecell{412 \\ \scriptsize{\textit{5.43\%}}}    & \makecell{606 \\ \scriptsize{\textit{7.99\%}}}     & \makecell{1,127 \\ \scriptsize{\textit{14.85\%}}} & 7,589   & 2 \\
        \textbf{UTKFace}\rule{0pt}{4ex}\rule[-2.5ex]{0pt}{0pt}  & \makecell{13,464 \\ \scriptsize{(\textit{55.85\%})}}  & \makecell{1,683 \\ \scriptsize{\textit{6.98\%}}}  & \makecell{2,749 \\ \scriptsize{\textit{11.40\%}}} & \makecell{1,683 \\ \scriptsize{\textit{6.98\%}}}  & \makecell{4,527 \\ \scriptsize{\textit{18.78\%}}} & 24,106  & 0 \\
        \textbf{MORPH}\rule{0pt}{4ex}\rule[-2.5ex]{0pt}{0pt}    & \makecell{40,069 \\ \scriptsize{(\textit{74.55\%})}}  & \makecell{5,011 \\ \scriptsize{\textit{9.32\%}}}  & \makecell{338 \\ \scriptsize{\textit{0.63\%}}}    & \makecell{4,992 \\ \scriptsize{\textit{9.29\%}}}  & \makecell{3,335 \\ \scriptsize{\textit{6.21\%}}}  & 53,745  & 1,863 \\
        \bottomrule
    \end{tabular}
    \caption{Split statistics per dataset. \textbf{Total} is the number of samples retained after filtering, partitioned into training, validation, and test subsets; \textbf{Disc.} is the number discarded, by the subject-age exclusivity algorithm or by face-detection failure. Validation and test splits divide into \textit{seen} classes (adults, 18--59) and \textit{unseen} classes (elderly 60+ at validation, children 0--17 at test)\footnote{Per-dataset filtering breakdowns are given in the supplementary material.}.}
    \label{tab:gzsl-benchmark-splits}
\end{table}

\subsection{Known Limitations of the Benchmark}
\label{sec:constraint-scope}

While we have selected the elderly group as a proxy for the unseen test set, that both groups require extrapolation does not establish that they require it in equal measure, nor that a model selected on one is optimally selected for the other. Facial appearance changes far more rapidly below 18 than above 60, and the cues separating a 6-year-old from a \mbox{12-year-old} have no clean analogue at the opposite extreme.

Furthermore, as we impose the constraint on children's data use for the age estimation we do not guarantee that previous steps followed suit. Every method we evaluate initializes from an ImageNet-pretrained backbone, as is the standard in computer vision.

ILSVRC-2012 contains no \texttt{person}, \texttt{child}, or \texttt{baby} category, which makes it tempting to treat pretraining as neutral. It is not. The 1000 categories include \texttt{bassinet}, \texttt{cradle}, \texttt{diaper}, and \texttt{bib} --- rarely photographed without an infant in frame --- alongside person-centric categories such as \texttt{bridegroom} and \texttt{military uniform}. \citet{yang2022faceobfuscation} observe that although ``most categories in the ImageNet challenge are not people categories,'' many incidental people appear, and annotate faces throughout the dataset for that reason; \citet{birhane2021pyrrhic} audited ILSVRC-2012 and found non-consensual imagery within it. A backbone pretrained on this corpus has in all likelihood seen children's faces. We therefore scope the contribution to the supervised age-estimation signal, not the derivation of prior representations. Extending it to pretraining --- an audited corpus, a from-scratch backbone, or self-supervision on verified data --- is a complement we do not attempt here.

\section{Experimental Results}

\subsection{Experimental Setup}

We evaluate nine methods on all six datasets, spanning the three paradigms: direct \textbf{Regression} and \textbf{DEX}~\cite{rothe2015dex}; the LDL family \textbf{DLDL}~\cite{gao2017dldl}, \textbf{DLDL-v2}~\cite{gao2018dldlv2}, \textbf{SORD}~\cite{diaz2019softlabels}, and \textbf{Mean-Var.}~\cite{pan2018meanvarianceloss}; and the rank-learning methods \textbf{OR-CNN}~\cite{niu2016multiplecnn}, \textbf{CORAL}~\cite{cao2020coral}, and \textbf{CORN}~\cite{shi2023corn}.

For all methods, we use a ResNet50~\cite{he2016resnet} pretrained on ImageNet~\cite{deng2009imagenet} as the backbone feature extractor --- the standard choice in this literature, and the reason our constraint applies to the age-estimation training signal rather than the full pipeline (Section~\ref{sec:constraint-scope}). Each method is equipped with its respective task-specific prediction head, and all heads are designed to operate over the age range of 0 to 101 years. For methods that model a probability distribution over the age space, we compute the predicted age as the median of the distribution instead of the mean, following \citet{paplhamcvpr2024reflect}, as the median is the estimator that minimizes MAE.

Each of the nine methods is evaluated on all six datasets under two experimental protocols. The first is a \textit{supervised} setting, which serves as a performance baseline: methods are trained and evaluated using the standardized data splits and subject-exclusivity constraints of \citet{paplhamcvpr2024reflect}. The second is the \textit{Generalized Zero-Shot Learning} setting defined by our proposed benchmark, where methods are trained exclusively on seen adult classes and evaluated on both seen and unseen age groups using our \mbox{benchmark splits.}

We run a dedicated hyperparameter search per method per setting, since paradigms differ in optimization dynamics --- one area where \citet{paplhamcvpr2024reflect} falls short, adopting a single configuration throughout. Learning rate and weight decay are tuned with Optuna~\cite{akiba2019optuna} (Tree-Structured Parzen Estimator) on UTKFace, optimizing validation MAE in the supervised setting and the harmonic mean of seen/unseen validation MAE under GZSL; search space, trial budget, optimizer, and batch size are identical across methods. Selected values span several orders of magnitude, confirming a shared configuration would disadvantage some paradigms\footnote{Details are in the supplementary material.}.

Across all experiments, training hyperparameters are held fixed, with only the learning rate and weight decay varying per method --- using the values found during hyperparameter search --- and the batch size varying per dataset, with smaller datasets receiving proportionally smaller batches. Rather than training for a fixed number of epochs, we fix the total number of training iterations to ensure a comparable computational budget across datasets of different sizes. Consistent with \citet{paplhamcvpr2024reflect}, all methods perform similarly under the supervised setting, with moderate MAE differences across methods, confirming that our individualized tuning procedure yields a fair and competitive baseline for each paradigm.

We then evaluate the same methods under the \textit{Generalized Zero-Shot Learning} setting using our benchmark. In these experiments, we report performance separately for the \textit{seen} (adults, 18--59) and \textit{unseen} (elderly, 60+ at validation; children, $<$18 at test) classes. We adopt three evaluation metrics: the MAE on seen classes, the MAE on unseen classes, and their harmonic mean, which we take as our primary metric following~\citet{xian2018zslthegoodbadugly}, as it penalizes methods that achieve strong performance on one class group at the expense of the other.

For all experiments and hyperparameter searches, images are preprocessed in a standardized manner, following a procedure similar to \citet{paplhamcvpr2024reflect}. We detect faces using RetinaFace, constructing a square bounding box from the detected facial landmarks. When multiple faces are present in an image, we retain the one with the largest area. The detected face region is then cropped and resized to \mbox{$256 \times 256$ pixels.}

\subsection{Results and Discussion}

We present results for all nine methods across both the supervised and GZSL settings. We first analyze performance under the supervised protocol as a baseline, then examine how methods behave when evaluated under our GZSL \mbox{benchmark.}

\begin{table}[t]
    \centering
    \renewcommand{\arraystretch}{1.05}
    \fontsize{8pt}{9.6pt}\selectfont 
    
    \setlength{\tabcolsep}{2pt}
    \begin{tabular}{lccccccc}
        \toprule
          & \textbf{AFAD} & \textbf{AgeDB} & \textbf{CACD} & \textbf{CLAP} & \textbf{UTKFace} & \textbf{MORPH} & \textit{\textbf{All}} \\
        \midrule
        \textbf{Regression} & 3.21 & \textbf{6.18} & \textbf{4.53} & \textbf{4.32} & \textbf{4.52} & \textbf{2.80} & \textbf{4.26} \\
        \textbf{DEX} & 3.34 & 7.16 & 4.69 & 6.19 & 5.25 & 3.38 & 5.00 \\
        \textbf{SORD} & 3.33 & 6.85 & 4.72 & 5.44 & 4.97 & 3.17 & 4.75 \\
        \textbf{DLDL} & 3.32 & 6.85 & 4.71 & 5.18 & 4.96 & 3.00 & 4.67 \\
        \textbf{DLDL-v2} & 3.34 & 6.54 & 4.66 & 5.12 & 4.73 & 3.00 & 4.56 \\
        \textbf{OR-CNN} & \textbf{3.17} & 6.31 & 4.54 & 4.71 & 4.57 & 2.86 & 4.36 \\
        \textbf{Mean-Var.} & \textbf{3.17} & 6.39 & 4.58 & 4.78 & 4.67 & 2.85 & 4.41 \\
        \textbf{CORAL} & 3.21 & 6.19 & 4.87 & 5.11 & 4.95 & 2.83 & 4.53 \\
        \textbf{CORN} & 3.33 & 6.73 & 4.57 & 5.39 & 5.04 & 2.98 & 4.67 \\
        \bottomrule
    \end{tabular}
    \caption{Test MAE across all datasets composing our benchmark, using the test splits proposed by \cite{paplhamcvpr2024reflect}, where all methods are trained under a common supervised protocol free of zero-shot constraints. CACD and CLAP in the column headers refer to CACD2000 and CLAP2016, respectively.}
    \label{tab:supervised-test-results}
\end{table}

In the supervised setting, Table~\ref{tab:supervised-test-results} reports test MAE for all methods across datasets under the supervised protocol. We can see that, for all methods, performance varies across datasets, reflecting differences in image acquisition conditions: more controlled datasets tend to yield lower errors, while in-the-wild datasets present greater variability. Nonetheless, methods achieve broadly comparable results, with aggregate MAE values averaging approximately 4 years. Among all methods, regression obtains the best aggregate performance and achieves the lowest MAE on all datasets except AFAD, where OR-CNN~\cite{niu2016multiplecnn} and Mean-Variance~\cite{pan2018meanvarianceloss} perform best.

\begin{table*}[t]
    \centering
    \renewcommand{\arraystretch}{1.15}
    \fontsize{8.4pt}{10.08pt}\selectfont 
    \setlength{\tabcolsep}{3pt}
    \begin{tabular}{lcccccccccccccccccccccccc}
        \toprule
          & \multicolumn{3}{c}{\textbf{AFAD}} & \multicolumn{3}{c}{\textbf{AgeDB}} & \multicolumn{3}{c}{\textbf{CACD2000}} & \multicolumn{3}{c}{\textbf{CLAP2016}} & \multicolumn{3}{c}{\textbf{UTKFace}} & \multicolumn{3}{c}{\textbf{MORPH}} & \multicolumn{3}{c}{\textit{\textbf{All}}} \\
        \cmidrule(lr{0pt}){2-4} \cmidrule(lr{0pt}){5-7} \cmidrule(lr{0pt}){8-10} \cmidrule(lr{0pt}){11-13} \cmidrule(lr{0pt}){14-16} \cmidrule(lr{0pt}){17-19} \cmidrule(lr{0pt}){20-22}
          & \textbf{S} & \textbf{U} & \textbf{H} & \textbf{S} & \textbf{U} & \textbf{H} & \textbf{S} & \textbf{U} & \textbf{H} & \textbf{S} & \textbf{U} & \textbf{H} & \textbf{S} & \textbf{U} & \textbf{H} & \textbf{S} & \textbf{U} & \textbf{H} & \textbf{S} & \textbf{U} & \textbf{H} \\
        \midrule
        \textbf{Regression} & 3.03 & 4.60 & 3.65 & 7.78 & 17.81 & 10.83 & \textbf{6.16} & 12.65 & 8.28 & \textbf{3.22} & 15.10 & \textbf{5.31} & \textbf{4.45} & 17.37 & \textbf{7.09} & 3.35 & 4.27 & 3.75 & \textbf{4.66} & 11.97 & \textbf{6.49} \\
        \textbf{DEX} & 3.07 & 4.79 & 3.74 & 8.64 & 19.15 & 11.90 & 6.28 & 12.55 & 8.37 & 4.52 & 15.26 & 6.98 & 5.19 & 18.94 & 8.15 & 3.07 & 3.04 & 3.05 & 5.13 & 12.29 & 7.03 \\
        \textbf{SORD} & 3.07 & 4.65 & 3.69 & 8.21 & 16.97 & 11.07 & 6.25 & \textbf{12.04} & \textbf{8.23} & 3.79 & 16.09 & 6.13 & 4.75 & 17.58 & 7.47 & 3.09 & \textbf{2.98} & 3.03 & 4.86 & 11.72 & 6.61 \\
        \textbf{DLDL} & 3.05 & 4.98 & 3.79 & 8.39 & 17.34 & 11.31 & 6.17 & 12.58 & 8.27 & 3.91 & 15.21 & 6.22 & 4.70 & 17.40 & 7.40 & 3.00 & 3.31 & 3.15 & 4.87 & 11.80 & 6.69 \\
        \textbf{DLDL-v2} & 3.12 & 4.76 & 3.77 & 8.08 & 17.29 & 11.01 & 6.30 & 13.45 & 8.58 & 3.85 & 16.31 & 6.22 & 4.60 & 18.18 & 7.34 & 2.96 & 3.35 & 3.15 & 4.82 & 12.23 & 6.68 \\
        \textbf{OR-CNN} & 3.21 & 4.81 & 3.85 & 7.60 & 17.42 & 10.58 & 6.34 & 14.79 & 8.88 & 3.73 & 14.63 & 5.95 & 4.61 & 17.86 & 7.32 & 3.03 & 3.74 & 3.35 & 4.75 & 12.21 & 6.65 \\
        \textbf{Mean-Var.} & \textbf{3.01} & \textbf{4.45} & \textbf{3.59} & 8.02 & \textbf{16.08} & 10.70 & 6.27 & 12.86 & 8.43 & 3.91 & \textbf{14.17} & 6.13 & 4.58 & \textbf{16.65} & 7.18 & 2.98 & 3.00 & \textbf{2.99} & 4.79 & \textbf{11.20} & 6.50 \\
        \textbf{CORAL} & 3.07 & 4.94 & 3.78 & 9.15 & 19.65 & 12.49 & 6.26 & 14.13 & 8.67 & 3.43 & 15.77 & 5.64 & 4.56 & 18.71 & 7.33 & \textbf{2.91} & 4.82 & 3.63 & 4.90 & 13.00 & 6.92 \\
        \textbf{CORN} \rule[-6pt]{0pt}{0pt} 
        & 3.21 & 4.80 & 3.84 & \textbf{7.33} & 18.09 & \textbf{10.43} & 6.21 & 13.74 & 8.55 & 4.05 & 15.38 & 6.41 & 4.89 & 17.53 & 7.64 & 3.07 & 3.52 & 3.28 & 4.79 & 12.18 & 6.69 \\
        \hline
        \textit{\textbf{Mean}} \rule[0pt]{0pt}{10pt} 
        & 3.09 & 4.75 & 3.74 & 8.13 & 17.76 & 11.15 & 6.25 & 13.20 & 8.47 & 3.82 & 15.32 & 6.11 & 4.70 & 17.80 & 7.44 & 3.05 & 3.56 & 3.26 & 4.84 & 12.07 & 6.70 \\
        \textit{\textbf{Std.}}  & 0.07 & 0.16 & 0.08 & 0.52 & 1.03 & 0.63 & 0.06 & 0.84 & 0.20 & 0.35 & 0.64 & 0.44 & 0.21 & 0.67 & 0.29 & 0.12 & 0.59 & 0.25 & 0.12 & 0.46 & 0.17 \\
        \bottomrule
    \end{tabular}
    \caption{Test MAE across all datasets on our GZSL benchmark, assessing how well age estimation models trained on adults generalize to children at test time. Metrics are reported for seen adult classes aged 18--59~(\textbf{S}), unseen children classes aged 0--17~(\textbf{U}), and their harmonic mean (\textbf{H}), which balances performance across both groups.}
    \label{tab:gzsl-test-results}
\end{table*}

Table~\ref{tab:gzsl-test-results} shows all methods performing comparably under GZSL, within datasets and in aggregate. On \textit{seen} classes they stay close to supervised levels ($4.84 \pm 0.12$ mean MAE). On \textit{unseen} ages every method degrades substantially and consistently ($12.07 \pm 0.46$). Taking the harmonic mean as the GZSL measure against supervised MAE, models perform $46.4\%$ worse on average\footnote{Per method, the percentage difference between its \textit{All} harmonic mean (Table~\ref{tab:gzsl-test-results}) and its \textit{All} supervised MAE (Table~\ref{tab:supervised-test-results}), averaged over the nine methods.}; at the extreme, CORAL on AgeDB more than doubles its supervised error (101.8\%).

These aggregates conceal a property that matters for reading the numbers: zero-shot difficulty is not constant, because the datasets do not cover the unseen interval equally. AFAD's 6,154 unseen test samples are \textit{all} aged 15--17 and MORPH's 3,335 span only 16--17, both adjacent to the seen boundary at 18, so a model need extrapolate barely a year. Unsurprisingly these show the mildest degradation (unseen MAE 4.75 and 3.56, against 17.80 on UTKFace, whose unseen split reaches infancy). The benchmark is therefore six protocols under one framework rather than a single task at one difficulty, and per-dataset errors should not be compared as though they measured the same thing. What generalizes is the \textit{direction} of the failure, not its magnitude.

A second caveat bounds precision rather than comparability. As noted in Table~\ref{tab:datasets}, most of these datasets do not carry ground-truth ages: AFAD infers them from upload dates, CACD2000 from birth years matched to search-engine timestamps, and CLAP2016 reports crowd-sourced \textit{apparent} age. Reported MAE therefore includes a dataset-dependent label-noise floor that we cannot separate from model error, and which is plausibly larger precisely where annotation was most indirect. This does not affect the finding --- a systematic 46.4\% degradation and a consistent anchoring pattern are far too large to attribute to labeling artifacts --- but it does mean the absolute error values should be read as indicative rather than exact.

\begin{figure*}
    \centering

    \begin{subfigure}[t]{0.155\linewidth}
        \centering
        \includegraphics[width=\linewidth]{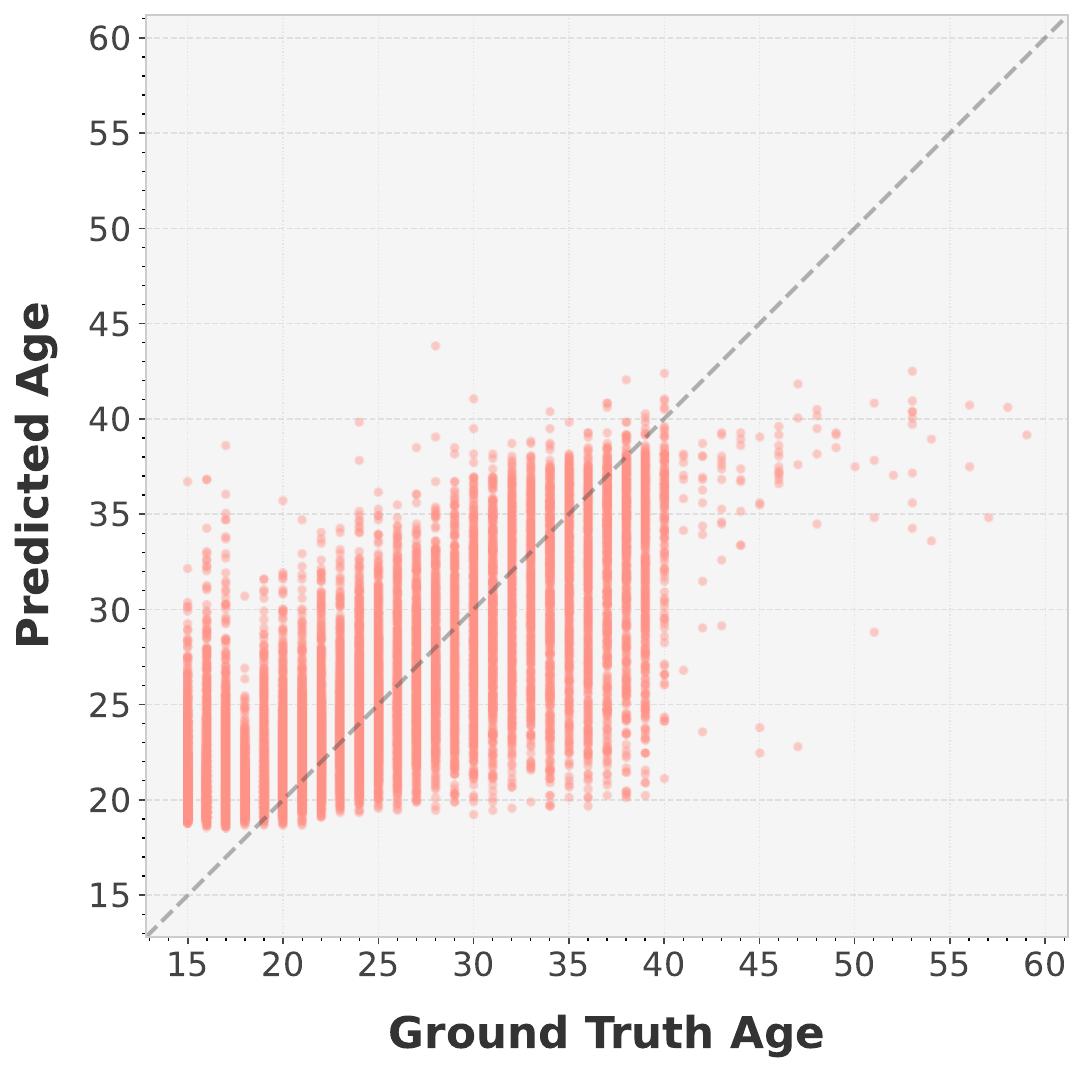}
        \caption{\footnotesize AFAD}
        \label{subfig:gzsl-afad-prediction-error}
    \end{subfigure}
    ~
    \begin{subfigure}[t]{0.155\linewidth}
        \centering
        \includegraphics[width=\linewidth]{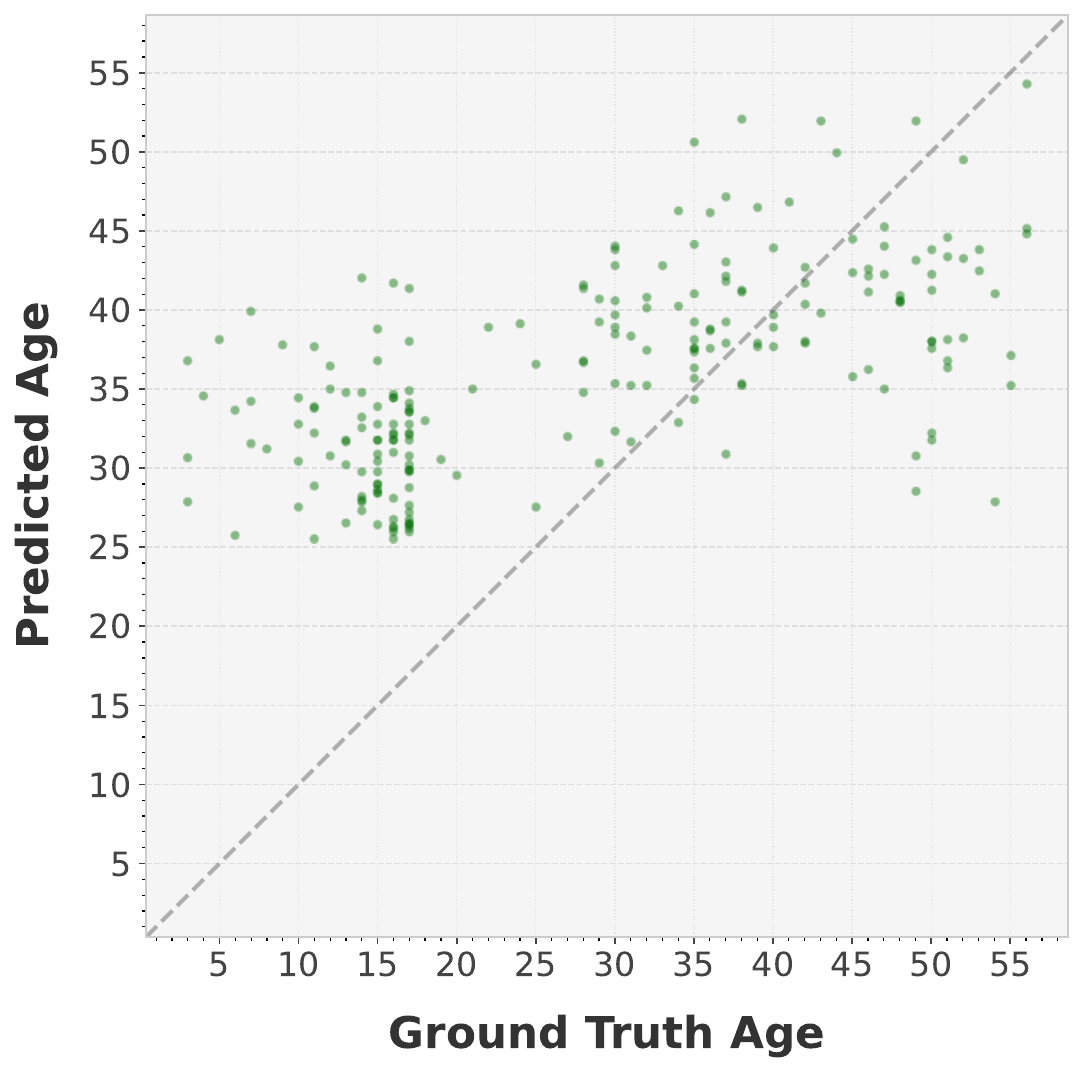}
        \caption{\footnotesize AgeDB}
        \label{subfig:gzsl-agedb-prediction-error}
    \end{subfigure}
    ~
    \begin{subfigure}[t]{0.155\linewidth}
        \centering
        \includegraphics[width=\linewidth]{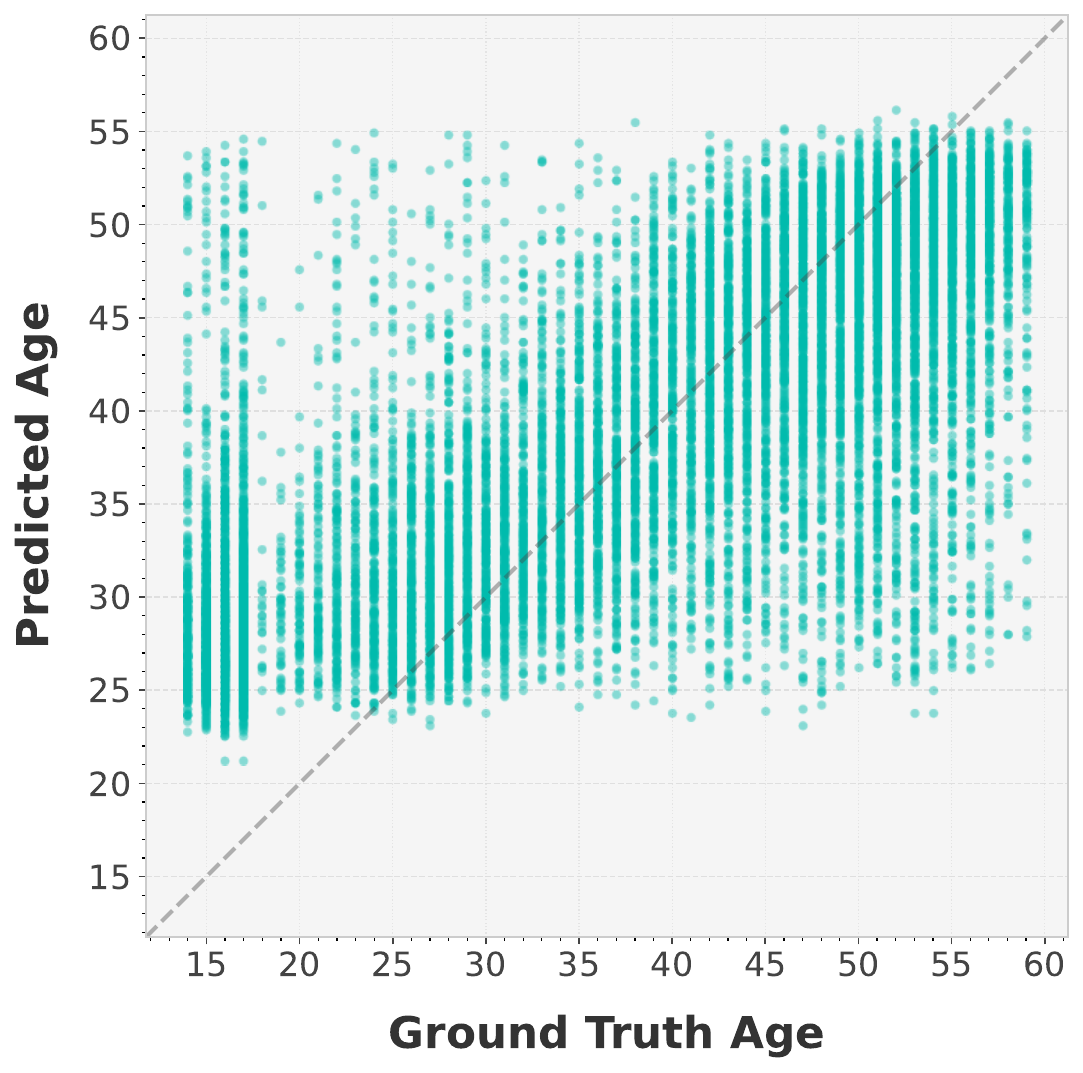}
        \caption{\footnotesize CACD2000}
        \label{subfig:gzsl-cacd2000-prediction-error}
    \end{subfigure}
    ~
    \begin{subfigure}[t]{0.155\linewidth}
        \centering
        \includegraphics[width=\linewidth]{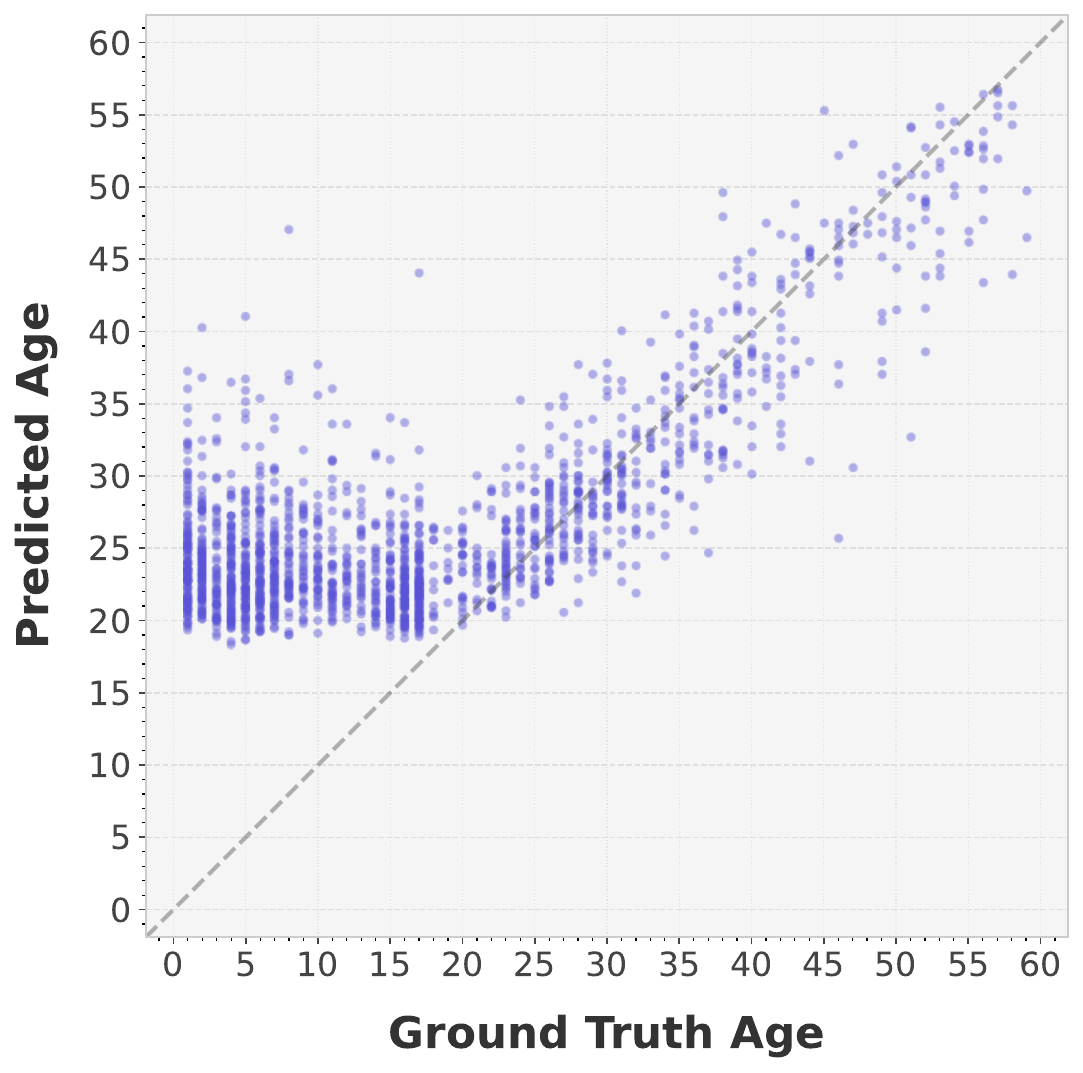}
        \caption{\footnotesize CLAP2016}
        \label{subfig:gzsl-clap2016-prediction-error}
    \end{subfigure}
    ~
    \begin{subfigure}[t]{0.155\linewidth}
        \centering
        \includegraphics[width=\linewidth]{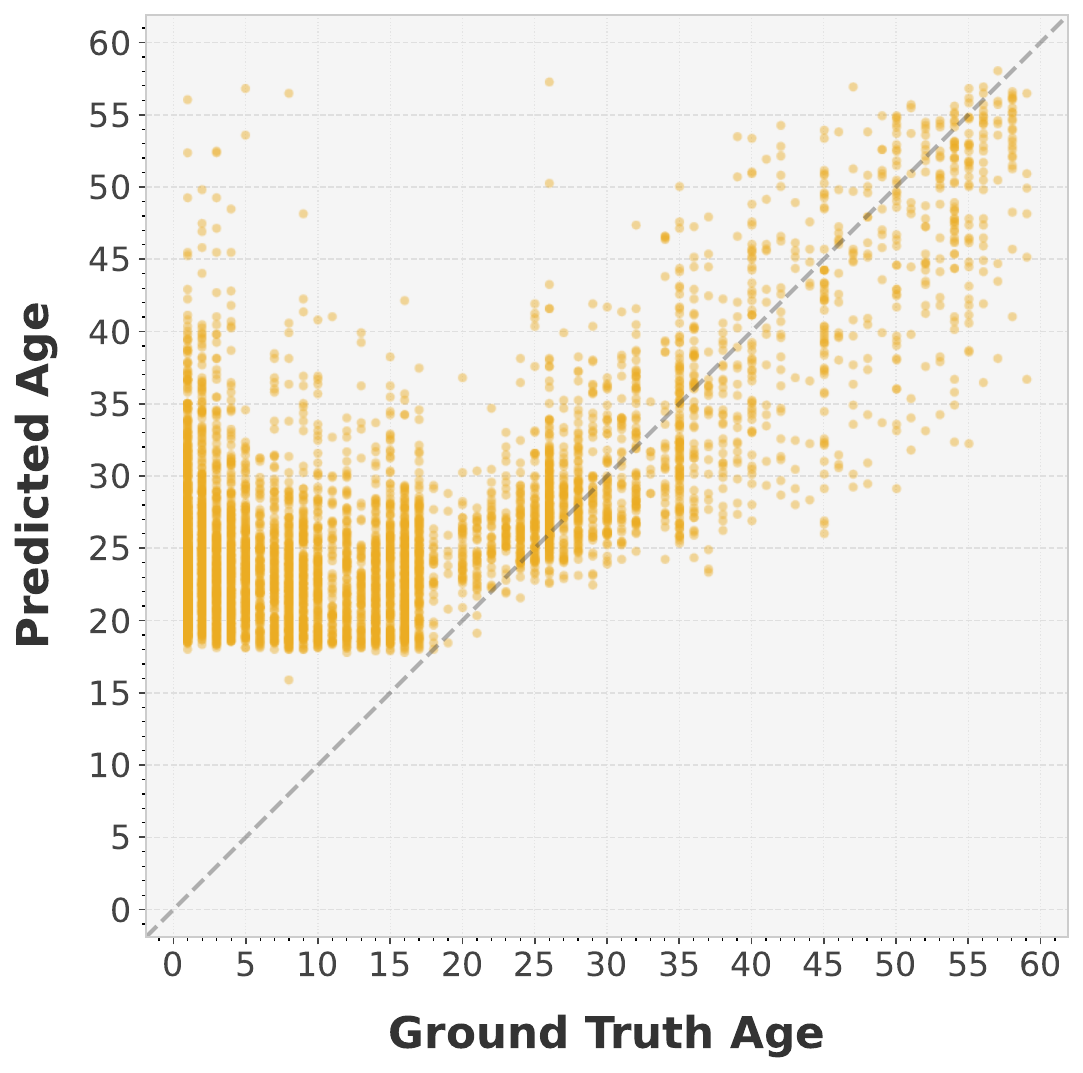}
        \caption{\footnotesize UTKFace}
        \label{subfig:gzsl-utkface-prediction-error}
    \end{subfigure}
    ~
    \begin{subfigure}[t]{0.155\linewidth}
        \centering
        \includegraphics[width=\linewidth]{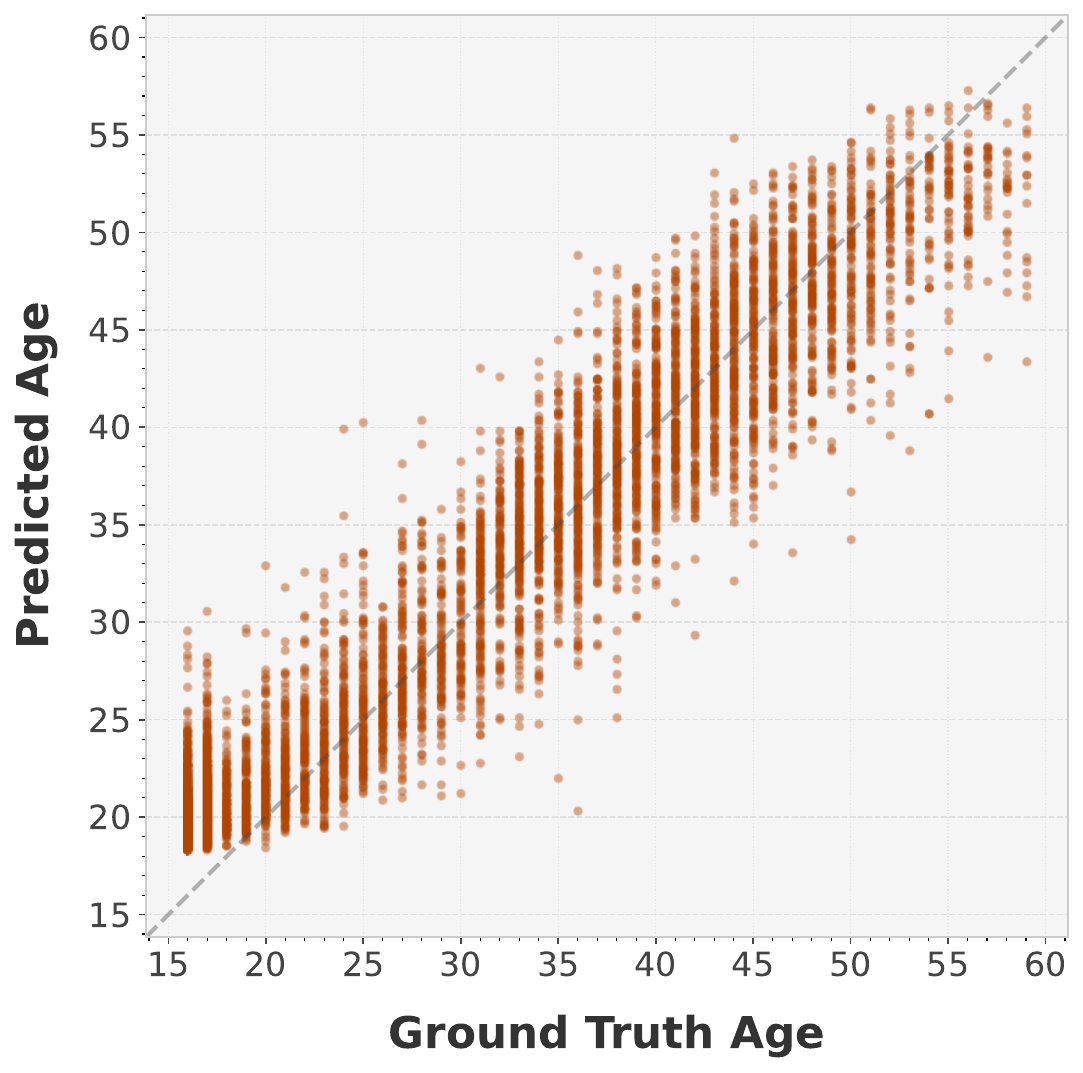}
        \caption{\footnotesize MORPH}
        \label{subfig:gzsl-morph-prediction-error}
    \end{subfigure}

    \caption{Scatter plots of \textbf{mean predicted age} \textit{vs} \textbf{ground-truth age} on the test split of our GZSL benchmark, shown separately for each of the six evaluation datasets. Each point represents a single test sample, where the predicted age is obtained by averaging the predictions of all evaluated methods for that sample. The dashed diagonal indicates perfect prediction, where predicted age is equal to ground-truth age; points above it correspond to over-estimated ages and points below to under-estimated~ages.}
    \label{fig:gzsl-results-aggregated-prediction-errors}

\end{figure*}

This pattern reflects a important challenge in GZSL: the tendency of models to be biased toward seen classes, systematically mapping unseen inputs to the nearest seen class rather than extrapolating to unobserved regions of the output space. Figure~\ref{fig:gzsl-results-aggregated-prediction-errors}, which aggregates prediction errors across all methods and datasets, makes this explicit: on average, models anchor their predictions for unseen ages to nearby seen classes, confirming that the learned representations fail to extrapolate beyond the training age distribution.

The mechanism differs by paradigm but the outcome does not. Classification methods such as DEX never encounter unseen ages as training targets, so the corresponding logits are optimized toward low probability and nothing induces mass there at inference. Rank-learning methods (OR-CNN, CORAL, CORN) fit each binary threshold classifier using seen samples only, so thresholds falling in unseen regions acquire degenerate decision rules. LDL methods are the interesting case: their smooth label distributions propagate supervision to neighboring ages, which ought in principle to transfer into adjacent unseen intervals. They do not. Supervision remains concentrated on observed ages, and while the predicted distribution is smoother than a one-hot output, its mass --- and therefore its median --- stays inside the \mbox{seen range.}

\section{Conclusion}

We introduced a generalized zero-shot benchmark for facial age estimation, motivated by a methodological gap and an ethical one. Current protocols evaluate under splits where training and test share an age distribution, so nothing in standard practice measures whether a model generalizes to an age interval it never saw. And children's images continue to be used in training without the safeguards that child-data governance frameworks call for~\citet{young2019rd4c} --- a practice the field extended rather than curtailed as it pursued accuracy on minors. We adopt the prohibition as a hard constraint and formalize the result as a generalized zero-shot problem: adults aged 18--59 are the seen classes; elderly individuals are the unseen validation classes, which permit model selection without touching children's images; children are the unseen test classes. Standardized subject-age-exclusive splits across six widely used datasets prevent both identity leakage and seen/unseen overlap, and an accompanying algorithm constructs them for datasets with identity annotations.

Nine state-of-the-art methods fail under this protocol. Seen-class MAE stays near supervised levels at $4.84 \pm 0.12$, while unseen-class MAE rises to $12.07 \pm 0.46$ --- a 46.4\% average degradation by harmonic mean, up to 52.8\% across methods and 101.8\% in the worst individual case. The failure spans classification, Label Distribution Learning, and rank learning alike, and it is structured rather than random: predictions for unseen ages are systematically anchored to nearby seen classes, the seen-class bias familiar from GZSL.

No method yet targets the setting our benchmark defines: joint seen/unseen evaluation, subject-age-exclusive splits, and a withheld interval fixed by a data-governance constraint rather than by data scarcity. The ordinal structure of the age label space is the obvious resource, supplying internally the auxiliary information that attribute-based ZSL imports from outside, and our nine baselines --- chosen to establish where the \textit{supervised} literature stands --- are the standard such a method would be measured against. A second is to evaluate zero-shot and vision-language models here. We deliberately did not: contemporary VLMs are pretrained on LAION-scale corpora documented to contain non-consensually scraped children's images~\cite{hrw2024brazilchildrenlaion, birhane2021multimodal}, so admitting them as baselines would import the very provenance problem the benchmark constrains.

The third addresses the data itself. Our benchmark reuses existing datasets and inherits their consent problems; a genuine remedy would be children's images collected ethically, most plausibly via informed consent from adults donating photographs of their own childhood. We did not attempt it. Such a collection is a standalone project requiring ethics-board approval, a governance and retention plan, and recruitment at scale. Most practically, a new dataset would not answer the question this paper asks --- whether existing methods, on the corpora the field uses, can operate under the constraint --- and would offer no basis for comparison with prior work. Ours adds no new collection and no new exposure and can be adopted immediately.

\section{Ethical Considerations}

\paragraph{Scope of the motivation.} We motivate this work through CSAI triage, where age estimation is an established proxy task and the target domain is inaccessible by law. The further constraint we impose --- excluding minors from training --- is our choice, adopted from child-data governance guidance rather than compelled by deployment. Neither justifies age verification or access control, currently the most active application of facial age estimation and one that subjects the very population this protocol protects to routine biometric assessment. A benchmark that makes age estimators easier to build serves that application whether or not we endorse it, and our constraint does nothing to prevent it.

\paragraph{Consent is not resolved by moving children to evaluation.} Restricting children's images to the test split reduces how far the model is shaped by them; it does not make their use consensual. The children in these datasets did not agree to appear in an age estimation benchmark, and neither did the adults, whose images were largely scraped. We identify no individuals, infer no further attributes, and add no personally identifiable information, and users of the benchmark should respect the source datasets' licenses --- but the data underneath it remains ethically encumbered.

\section{Acknowledgments}

This work is partially funded by FAPESP \mbox{2023/12086-9}. C.~Petrucci is also funded by FAPESP \mbox{2024/09375-1}. S.~Avila is also funded by FAPESP 2023/12865-8, \mbox{2020/09838-0}, 2013/08293-7, CNPq 316489/2023-9, and H.IAAC 01245.003479/2024-10.

\bibliography{aaai2027}


\section{Supplementary Material}
\appendix
\setcounter{secnumdepth}{1} 
\setcounter{figure}{0}
\setcounter{table}{0}
\setcounter{equation}{0}
\setcounter{algorithm}{0}
\renewcommand{\thefigure}{A\arabic{figure}}
\renewcommand{\thetable}{A\arabic{table}}
\renewcommand{\theequation}{A\arabic{equation}}
\renewcommand{\thealgorithm}{A\arabic{algorithm}}
\section{Split Construction Algorithm}

Algorithm~\ref{alg:gzsl-split-construction-subject-age-exclusivity} assigns every image of a dataset to one of five \textsc{folders} while enforcing subject-age exclusivity.

\begin{algorithm*}[t]
    \caption{GZSL split construction with subject-age exclusivity.}
    \label{alg:gzsl-split-construction-subject-age-exclusivity}
    \textbf{Input}: Annotations $\mathcal{D} = \{(s_i, a_i, x_i)\}$ of subject IDs, ages, and images \\
    \textbf{Parameter}: Adult range $[a_\text{min}, a_\text{max})$; adult split fractions $\alpha$ (train), $\beta$ (val) \\
    \textbf{Output}: \textsc{Folder} assignment $f : \mathcal{D} \to \{0, 1, 2, 3, 4\}$, where \textsc{folders} 0--2 hold seen train/val/test and \textsc{folders} 3--4 hold unseen val/test
    \begin{algorithmic}[1]
        \STATE Partition each subject's images into $\mathcal{M}_s$, $\mathcal{A}_s$, $\mathcal{E}_s$ (minors, adults, elders); classify subjects as \textit{minor-only}, \textit{adult-only}, \textit{elder-only}, or \textit{mixed}
        \STATE Compute \textsc{folder} targets $t_k$ from dataset totals; initialise running counts $r_k$ with contributions of non-mixed subjects (adult-only subjects assumed to follow $\alpha / \beta / (1{-}\alpha{-}\beta)$)
        \FORALL{mixed subjects $s$}
            \FORALL{candidate assignments $c = (\text{split},\; \text{kept group})$}
                \STATE Let $k_c$ be the \textsc{folder} implied by $c$ and $n_c = |\text{images of } s \text{ in kept group}|$
                \STATE $\mathrm{score}(c) \gets n_c \cdot \max(0,\; t_{k_c} - r_{k_c}) \;/\; t_{k_c}^{\,2}$
            \ENDFOR
            \STATE $c^\star \gets \arg\max_c\; \mathrm{score}(c)$ \COMMENT{tie-break: test $\succ$ val $\succ$ train}
            \STATE Assign $s$ to \textsc{folder} $k_{c^\star}$; update $r_{k_{c^\star}} \mathrel{+}= n_{c^\star}$; discard images of $s$ outside the kept group of $c^\star$
        \ENDFOR
        \FOR{each unseen \textsc{folder} $k \in \{3, 4\}$ with $r_k < t_k$}
            \STATE Reroute unassigned mixed subjects, largest first, until $r_k \geq t_k$
        \ENDFOR
        \STATE Sort adult-only subjects by $|\mathcal{A}_s|$ descending; greedily fill \textsc{folders} 0, 1, 2 to meet the residual $\alpha / \beta / (1{-}\alpha{-}\beta)$ target
        \RETURN $f$, mapping each retained image to its assigned \textsc{folder}
    \end{algorithmic}
\end{algorithm*}

\subsubsection{Target Counts}

Before assignment begins, we compute a \textit{target count} $t_k$ for each \textsc{folder}, defined as the number of images the \textsc{folder} would hold under an ideal partition where no subject-exclusivity constraint applied --- that is, if every image of every subject could be freely placed into the \textsc{folder} corresponding to its age group. These targets serve as reference quantities that the algorithm seeks to approximate while respecting subject-age exclusivity.

\subsubsection{Candidate Scoring}

Each subject is classified into one of four types according to which age groups its images cover: \textit{minor-only}, \textit{adult-only}, \textit{elder-only}, or \textit{mixed}. Single-group subjects are assigned to their corresponding \textsc{folder} without ambiguity. Mixed subjects, whose images span at least two age groups, require an explicit decision: for each such subject, we enumerate candidate assignments $c = (\text{split},\; \text{kept age group})$ --- the target \textsc{folder} being fully determined by this pair --- and select the one that maximizes
\begin{equation}
    \mathrm{score}(c) = n_c \cdot \frac{\max(0,\; t_c - r_c)}{t_c^2},
    \label{eq:gzsl-candidate-score}
\end{equation}
where $n_c$ is the number of images retained under candidate assignment $c$, $t_c$ is the target count of the implied \textsc{folder}, and $r_c$ is its running count, accounting for contributions already fixed by non-mixed subjects and by mixed subjects assigned in previous iterations.

The numerator $\max(0,\; t_c - r_c)$ measures the \textsc{folder}'s \textit{remaining demand}, while dividing by $t_c^2$ normalizes urgency proportionally to \textsc{folder} size, prioritizing small unseen \textsc{folders} (e.g., \textsc{folder}~3, which must hold all minor images in the dataset) over the much larger training \textsc{folder}. Ties are broken in favor of test over validation over training, and the images of the subject that fall outside the selected age group are discarded.

\subsubsection{Correction Sweep}

Following this greedy pass, a correction sweep reroutes subjects, in descending order of contribution, to fill any unseen \textsc{folder} that remains below its target. Finally, adult-only subjects are distributed across the seen training, validation, and testing \textsc{folders} in an $80\%/10\%/10\%$ ratio by image count.

\subsubsection{Reproducibility} The procedure is deterministic. Assignments for single-group subjects, and all mixed-subject decisions, depend only on the annotation order and the running counts. The one stochastic step, the ordering of adult-only subjects when filling the seen \textsc{folders}, is seeded, so running the assignment on the released annotations reproduces the published partitions exactly.

\section{Age Estimation Datasets}

The benchmark is assembled from six facial age estimation datasets:

\begin{itemize}
    \item \textbf{AFAD}~\cite{niu2016multiplecnn} contains 165,501 facial images collected from profile pictures of users from the RenRen social network. Age labels are derived automatically by computing the difference between each user's date of birth and the date their photo was uploaded. It focuses predominantly on East Asian individuals and covers ages from 15 to 40 years old, making it particularly useful for studying age estimation under a more demographically constrained setting.

    \item \textbf{AgeDB}~\cite{moschoglou2017agedb} is an \textit{in-the-wild} dataset composed of 16,488 images from 568 individuals, including actors, politicians, scientists, and other public figures. Its images cover ages ranging from 1 to 101 years old, and are manually annotated to provide relatively accurate age labels across large temporal variations.

    \item \textbf{CACD2000}~\cite{chunchen2014cacd} contains 163,446 images from 2,000 celebrities collected from the Internet, covering ages from 16 to 62 years old. Since the images were retrieved through search engines using celebrity names and timestamps, age labels were estimated from birth years and image dates, introducing potential annotation inaccuracies.

    \item \textbf{CLAP2016}~\cite{agustsson2017apparentrealage}, or APPA-REAL, contains 7,591 facial images annotated with both real and apparent age labels. It was collected for the ChaLearn Looking at People Challenge and includes approximately 7,000 subjects ranging from 0 to 95~years old, with apparent ages obtained via crowd-sourced annotations.

    \item \textbf{UTKFace}~\cite{zhang2017utkface} contains 24,106 \textit{in-the-wild} facial images annotated with age, gender, and ethnicity labels, covering ages from 0 to 116 years old. It includes large variations in pose, illumination, ethnicity, and facial appearance. Some labels are estimated rather than manually verified.

    \item \textbf{MORPH}~\cite{ricanek2006morph} is one of the most widely used datasets in facial age estimation. It contains 55,608 images from 13,617 subjects between 16 and 77 years old. However, the dataset consists primarily of mug-shot style images of incarcerated individuals, introducing social and demographic characteristics that differ from unconstrained facial datasets.
\end{itemize}

\begin{figure*}
    \centering

    \begin{subfigure}[t]{0.155\linewidth}

        \setlength{\tabcolsep}{-0.1pt}  
        \renewcommand{\arraystretch}{0}  

        \begin{tabular}{cc}
            \begin{minipage}[b]{0.535\linewidth}
                \centering
                \includegraphics[width=\linewidth,max width=\linewidth, max height=\linewidth,keepaspectratio]{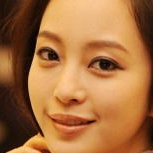}
            \end{minipage} &
            \begin{minipage}[b]{0.535\linewidth}
                \centering
                \includegraphics[width=\linewidth,max width=\linewidth, max height=\linewidth,keepaspectratio]{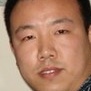}
            \end{minipage} \\
            \begin{minipage}[b]{0.535\linewidth}
                \centering
                \includegraphics[width=\linewidth,max width=\linewidth, max height=\linewidth,keepaspectratio]{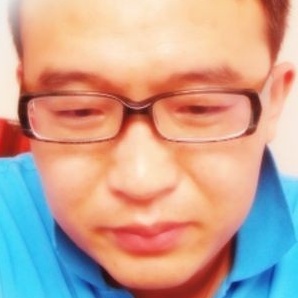}
            \end{minipage} &
            \begin{minipage}[b]{0.535\linewidth}
                \centering
                \includegraphics[width=\linewidth,max width=\linewidth, max height=\linewidth,keepaspectratio]{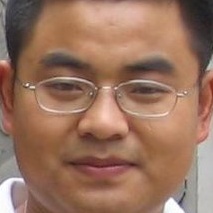}
            \end{minipage} \\
        \end{tabular}

        \caption{\footnotesize AFAD}
        \label{subfig:afad-samples}

    \end{subfigure}
    ~
    \begin{subfigure}[t]{0.155\linewidth}

        \setlength{\tabcolsep}{-0.1pt}  
        \renewcommand{\arraystretch}{0}  

        \begin{tabular}{cc}
            \begin{minipage}[b]{0.535\linewidth}
                \centering
                \includegraphics[width=\linewidth,max width=\linewidth, max height=\linewidth,keepaspectratio]{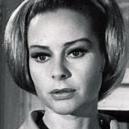}
            \end{minipage} &
            \begin{minipage}[b]{0.535\linewidth}
                \centering
                \includegraphics[width=\linewidth,max width=\linewidth, max height=\linewidth,keepaspectratio]{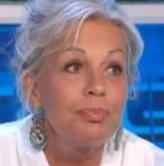}
            \end{minipage} \\
            \begin{minipage}[b]{0.535\linewidth}
                \centering
                \includegraphics[width=\linewidth,max width=\linewidth, max height=\linewidth,keepaspectratio]{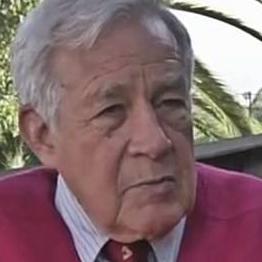}
            \end{minipage} &
            \begin{minipage}[b]{0.535\linewidth}
                \centering
                \includegraphics[width=\linewidth,max width=\linewidth, max height=\linewidth,keepaspectratio]{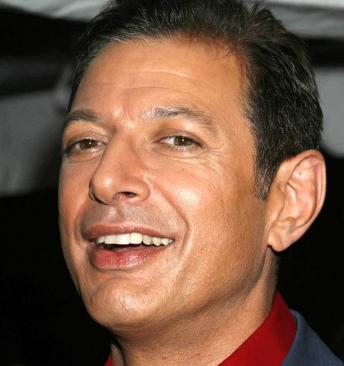}
            \end{minipage} \\
        \end{tabular}

        \caption{\footnotesize AgeDB}
        \label{subfig:agedb-samples}

    \end{subfigure}
    ~
    \begin{subfigure}[t]{0.155\linewidth}

        \setlength{\tabcolsep}{-0.1pt}  
        \renewcommand{\arraystretch}{0}  

        \begin{tabular}{cc}
            \begin{minipage}[b]{0.535\linewidth}
                \centering
                \includegraphics[width=\linewidth,max width=\linewidth, max height=\linewidth,keepaspectratio]{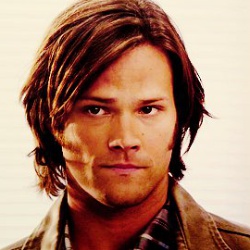}
            \end{minipage} &
            \begin{minipage}[b]{0.535\linewidth}
                \centering
                \includegraphics[width=\linewidth,max width=\linewidth, max height=\linewidth,keepaspectratio]{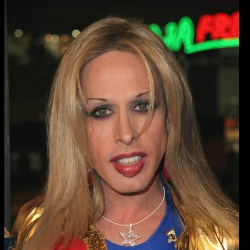}
            \end{minipage} \\
            \begin{minipage}[b]{0.535\linewidth}
                \centering
                \includegraphics[width=\linewidth,max width=\linewidth, max height=\linewidth,keepaspectratio]{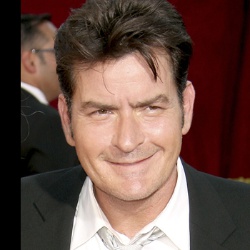}
            \end{minipage} &
            \begin{minipage}[b]{0.535\linewidth}
                \centering
                \includegraphics[width=\linewidth,max width=\linewidth, max height=\linewidth,keepaspectratio]{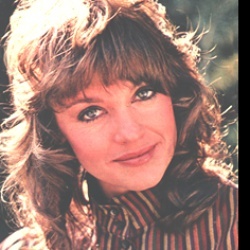}
            \end{minipage} \\
        \end{tabular}

        \caption{\footnotesize CACD2000}
        \label{subfig:cacd2000-samples}

    \end{subfigure}
    ~
    \begin{subfigure}[t]{0.155\linewidth}

        \setlength{\tabcolsep}{-0.1pt}  
        \renewcommand{\arraystretch}{0}  

        \begin{tabular}{cc}
            \begin{minipage}[b]{0.535\linewidth}
                \centering
                \includegraphics[width=\linewidth,max width=\linewidth, max height=\linewidth,keepaspectratio]{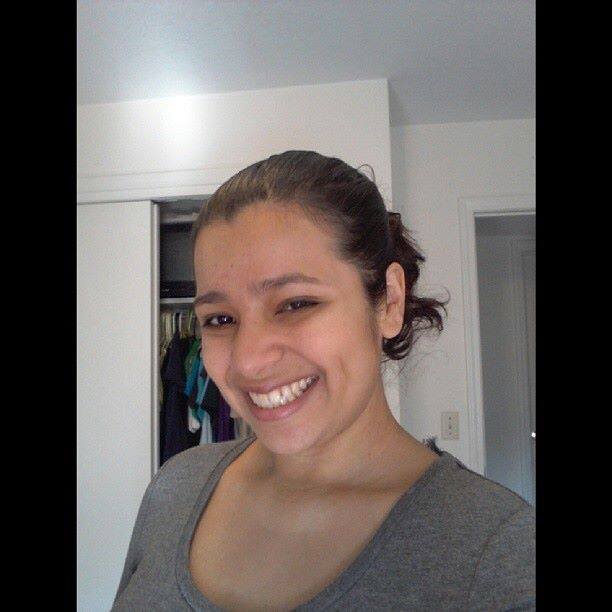}
            \end{minipage} &
            \begin{minipage}[b]{0.535\linewidth}
                \centering
                \includegraphics[width=\linewidth,max width=\linewidth, max height=\linewidth,keepaspectratio]{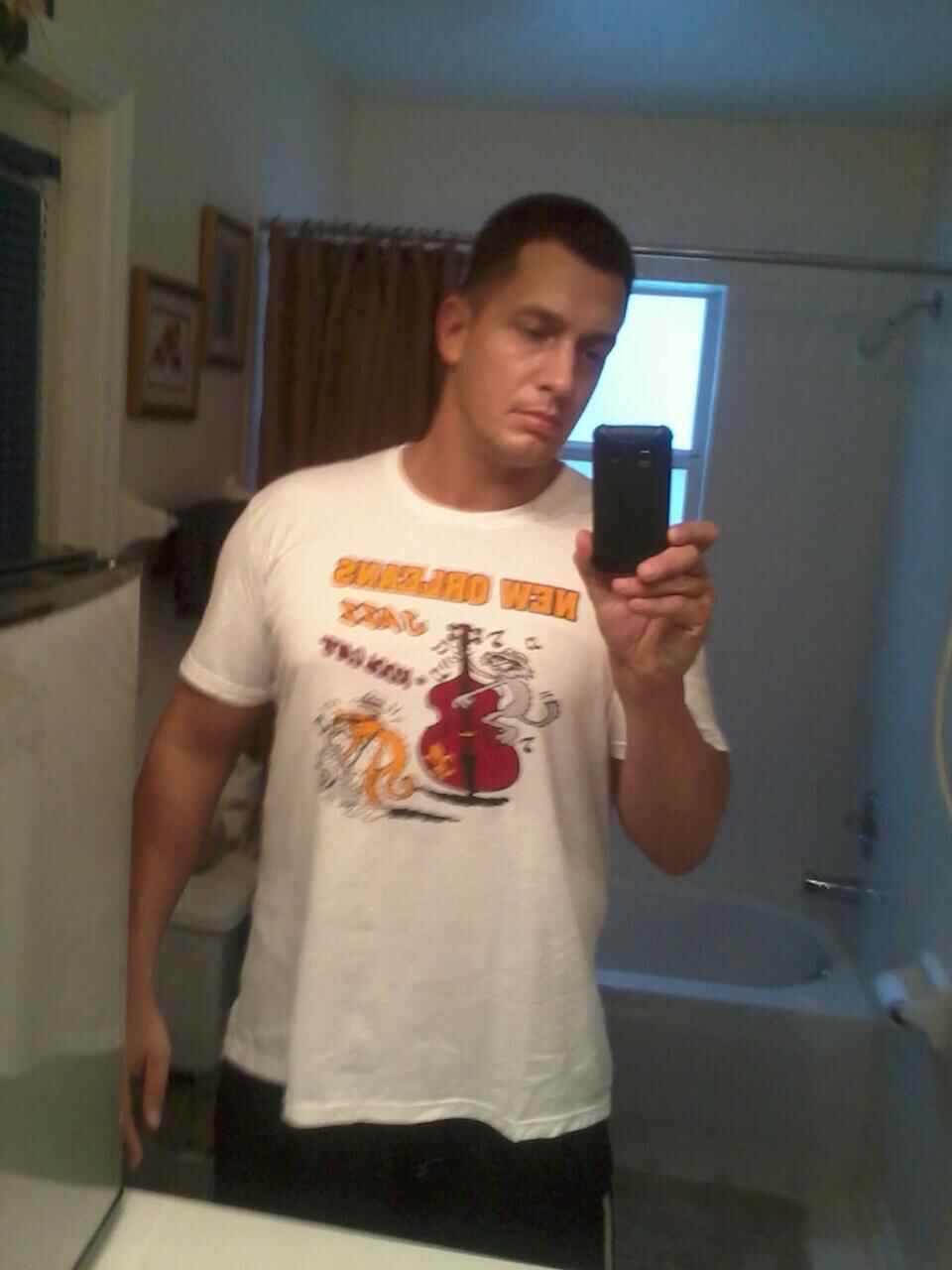}
            \end{minipage} \\
            \begin{minipage}[b]{0.535\linewidth}
                \centering
                \includegraphics[width=\linewidth,max width=\linewidth, max height=\linewidth,keepaspectratio]{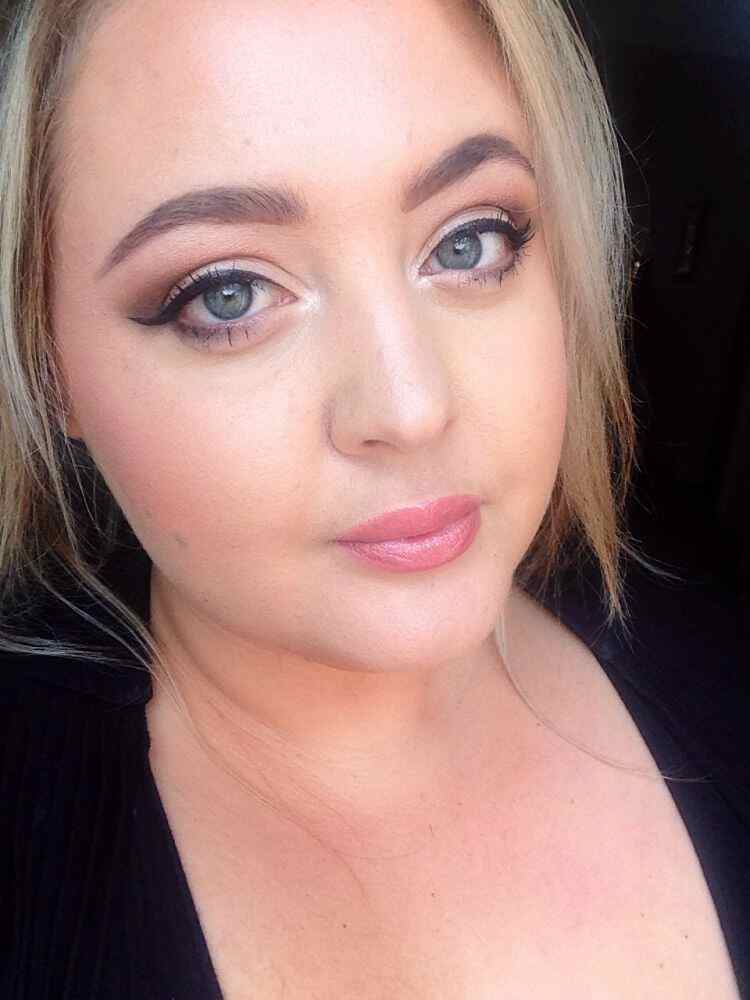}
            \end{minipage} &
            \begin{minipage}[b]{0.535\linewidth}
                \centering
                \includegraphics[width=\linewidth,max width=\linewidth, max height=\linewidth,keepaspectratio]{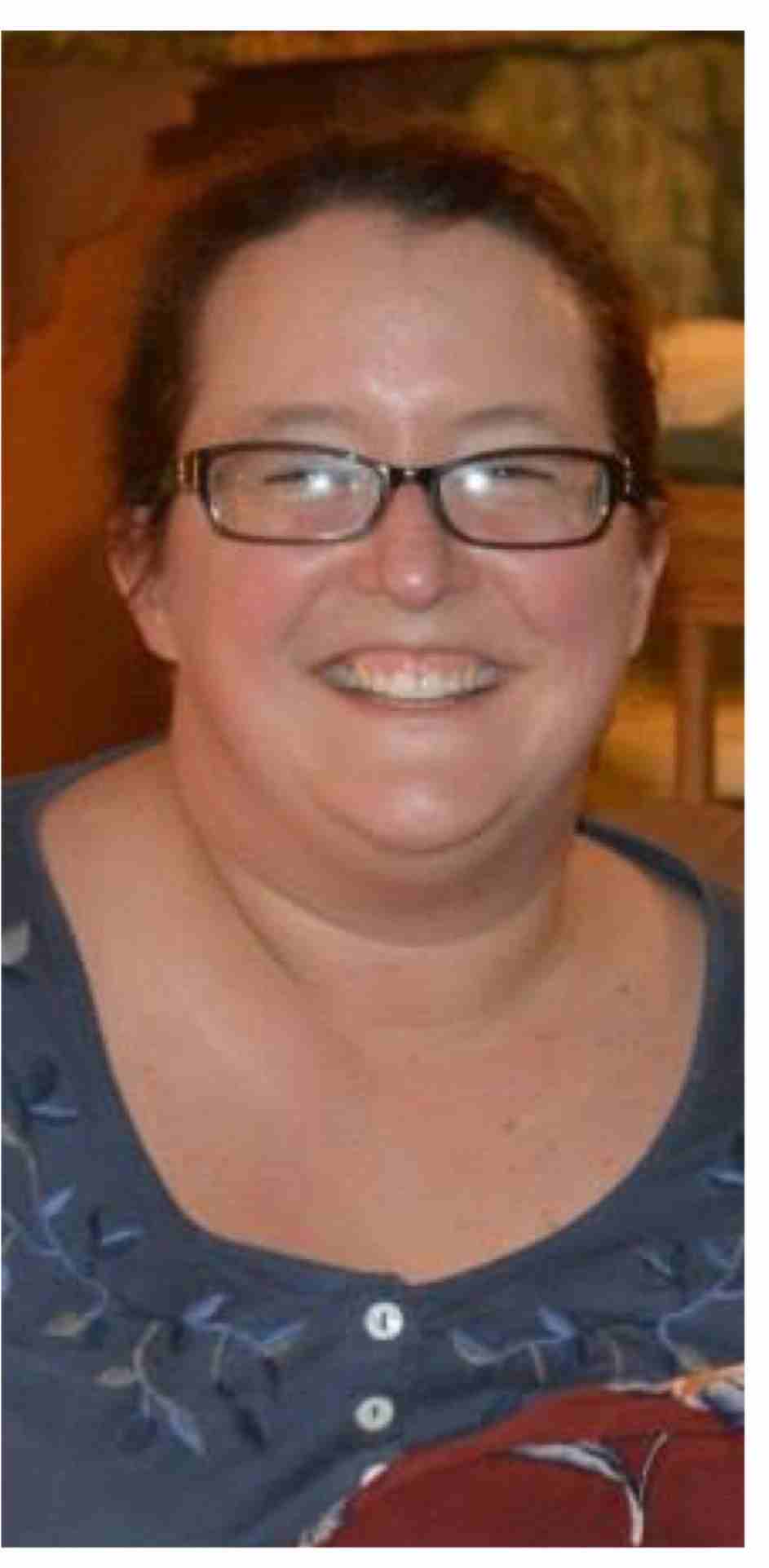}
            \end{minipage} \\
        \end{tabular}

        \caption{\footnotesize CLAP2016}
        \label{subfig:clap2016-samples}

    \end{subfigure}
    ~
    \begin{subfigure}[t]{0.155\linewidth}

        \setlength{\tabcolsep}{-0.1pt}  
        \renewcommand{\arraystretch}{0}  

        \begin{tabular}{cc}
            \begin{minipage}[b]{0.535\linewidth}
                \centering
                \includegraphics[width=\linewidth,max width=\linewidth, max height=\linewidth,keepaspectratio]{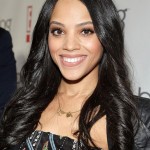}
            \end{minipage} &
            \begin{minipage}[b]{0.535\linewidth}
                \centering
                \includegraphics[width=\linewidth,max width=\linewidth, max height=\linewidth,keepaspectratio]{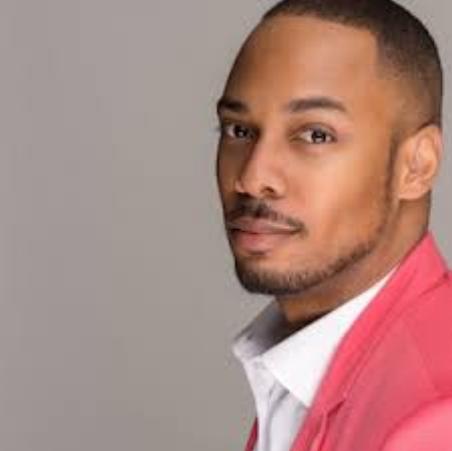}
            \end{minipage} \\
            \begin{minipage}[b]{0.535\linewidth}
                \centering
                \includegraphics[width=\linewidth,max width=\linewidth, max height=\linewidth,keepaspectratio]{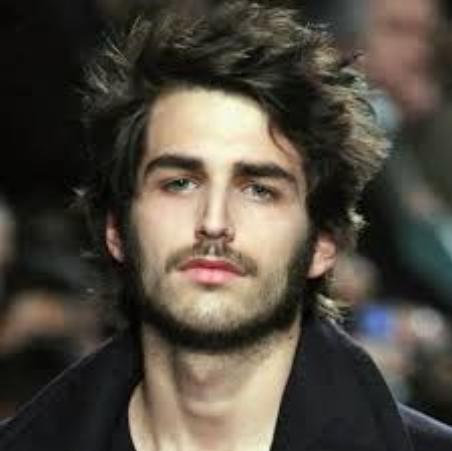}
            \end{minipage} &
            \begin{minipage}[b]{0.535\linewidth}
                \centering
                \includegraphics[width=\linewidth,max width=\linewidth, max height=\linewidth,keepaspectratio]{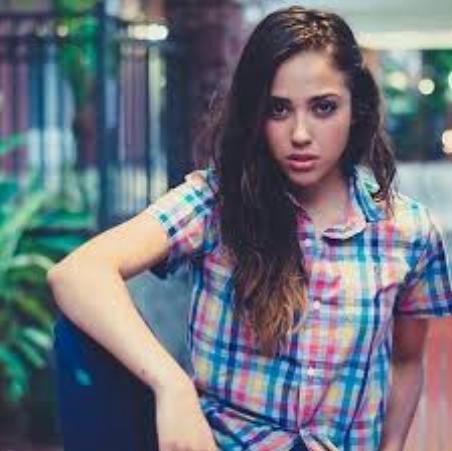}
            \end{minipage} \\
        \end{tabular}
        \caption{\footnotesize UTKFace}
        \label{subfig:utkface-samples}

    \end{subfigure}
    ~
    \begin{subfigure}[t]{0.155\linewidth}
        \setlength{\tabcolsep}{-0.1pt}  
        \renewcommand{\arraystretch}{0}  
        \begin{tabular}{cc}
            \begin{minipage}[b]{0.535\linewidth}
                \centering
                \includegraphics[width=\linewidth,max width=\linewidth, max height=\linewidth,keepaspectratio]{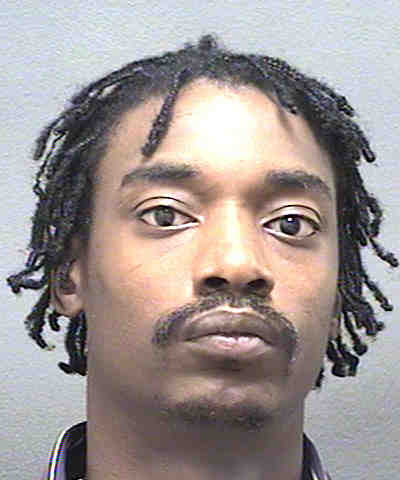}
            \end{minipage} &
            \begin{minipage}[b]{0.535\linewidth}
                \centering
                \includegraphics[width=\linewidth,max width=\linewidth, max height=\linewidth,keepaspectratio]{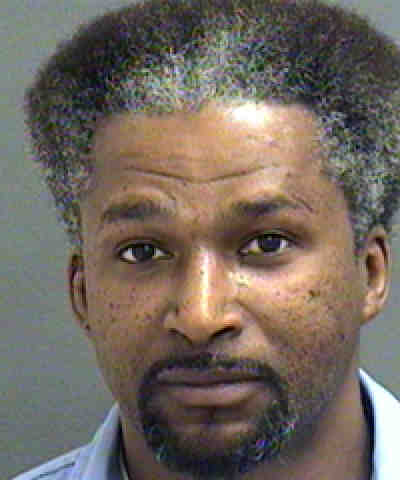}
            \end{minipage} \\
            \begin{minipage}[b]{0.535\linewidth}
                \centering
                \includegraphics[width=\linewidth,max width=\linewidth, max height=\linewidth,keepaspectratio]{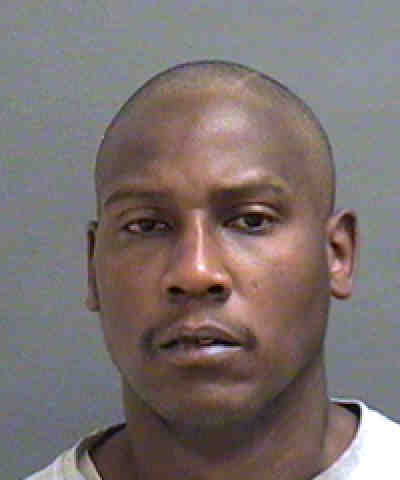}
            \end{minipage} &
            \begin{minipage}[b]{0.535\linewidth}
                \centering
                \includegraphics[width=\linewidth,max width=\linewidth, max height=\linewidth,keepaspectratio]{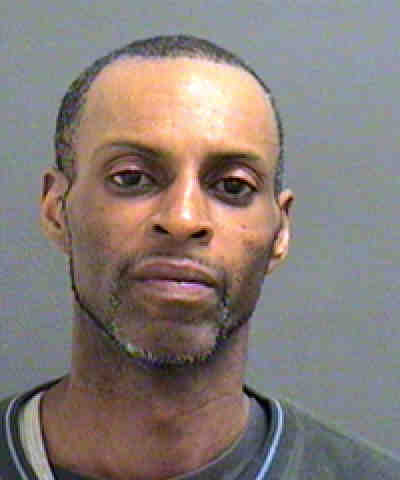}
            \end{minipage} \\
        \end{tabular}

        \caption{\footnotesize MORPH}
        \label{subfig:morph-samples}

    \end{subfigure}

    \caption{Randomly sampled images from each of the six datasets composing our benchmark, illustrated as $2 \times 2$ grids. Each grid corresponds to one dataset, ordered left to right as AFAD, AgeDB, CACD2000, CLAP2016, UTKFace, and MORPH.}
    \label{fig:dataset-samples}

\end{figure*}

Overall, these datasets differ substantially not only in demographic composition and image conditions but also in collection procedures and annotation reliability. For instance, AFAD predominantly represents East Asian individuals, whereas CACD2000 and AgeDB are strongly influenced by celebrity imagery. Similarly, MORPH presents a particularly distinct social context due to its mug-shot acquisition protocol. Such differences reinforce the importance of standardized and carefully controlled evaluation protocols when comparing methods across datasets.

\section{Sample Filtering Statistics}

As shown in Table~\ref{tab:supp-gzsl-sample-filtering}, the number of samples discarded during filtering varies considerably across datasets. AFAD, CLAP2016, and UTKFace require no exclusivity filtering, so no samples are lost: CLAP2016 and UTKFace carry no subject annotations, and AFAD's subjects fall within a narrow age range, making them naturally compatible with the benchmark constraints.

In contrast, CACD2000 loses 20,543 images (12.6\% of available samples) despite being a large dataset: with only 2,000 celebrity subjects and an average of approximately 81.7 images each, many individuals span a wide range of ages, forcing the algorithm to discard the images that conflict with each subject's assigned split. Alternatively, MORPH loses comparatively few images (1,613, or 2.9\%): although it is a longitudinal dataset, its child portion covers only ages 16--17 and its elderly representation is sparse, so cross-group overlap remains limited.

\begin{table}[t]
    \centering
    \footnotesize
    \renewcommand{\arraystretch}{1.35}
    \setlength{\tabcolsep}{5pt}
    \begin{tabular}{lrrrr}
        \toprule
         & \multirow{2}{*}{\textbf{Available}} & \multicolumn{2}{c}{\textbf{Filtered}} & \multirow{2}{*}{\textbf{Selected}} \\
        \cmidrule(lr{0pt}){3-4}
         & & \textbf{Exclusivity} & \textbf{No faces} & \\
        \midrule
        \textbf{AFAD}     & 165,501 & 0      & 46  & 165,455 \\
        \textbf{AgeDB}    & 16,488  & 10,814 & 1   & 5,673 \\
        \textbf{CACD2000} & 163,446 & 20,543 & 4   & 142,899 \\
        \textbf{CLAP2016} & 7,591    & 0      & 2   & 7,589 \\
        \textbf{UTKFace}  & 24,106  & 0      & 0   & 24,106 \\
        \textbf{MORPH}    & 55,608  & 1,613  & 250 & 53,745 \\
        \bottomrule
    \end{tabular}
    \caption{Sample filtering statistics per dataset. Starting from the publicly available splits (\textbf{Available}), samples are filtered either by our subject- and age-exclusivity algorithm (\textbf{Exclusivity}) or by preprocessing failures such as corrupted images or undetected faces (\textbf{No faces}). \textbf{Selected} reports the number of samples kept after both filtering stages and used to compose our GZSL benchmark.}
    \label{tab:supp-gzsl-sample-filtering}
\end{table}

AgeDB is by far the most severely affected, with 10,814 discarded images (65.6\%). With only 568 subjects across 16,488 images --- an average of 29.0 per subject --- and an age range spanning from 0 to 101 years, many individuals appear across multiple age groups that span the seen/unseen boundary, forcing the algorithm to discard the majority of each such subject's images. As a result, 70.83\% of AgeDB's retained samples fall in the unseen validation set, leaving only 1,270 images (22.39\%) for training --- making it the most constrained dataset in the benchmark.

Two cases stand out for their particularly small unseen validation sets. AFAD, despite its large total of 165,455 samples, contributes only 154 images (0.09\%) to the unseen validation partition, since the dataset consists predominantly of university students and young adults with very few individuals above 60. MORPH tells a similar story: its unseen validation set holds just 338 samples (0.63\% of 53,745 total), reflecting an age distribution concentrated in the adult range with sparse elderly coverage.

\section{Composition of the Unseen Test Split}

The benchmark withholds every sample under 18 from training and evaluates on it at test time. The age distribution of that withheld population varies considerably across the six datasets, which affects how the per-dataset results should \mbox{be read.}

Table~\ref{tab:test-split-composition} reports the age composition of the unseen test split (\textsc{folder}~3) for each dataset. In AFAD, CACD2000 and MORPH the withheld samples are all adolescents, covering ages 15--17, 14--17 and 16--17 respectively; none of these datasets contains a child under 13. UTKFace and CLAP2016 are the only two with a large proportion of young children, at 57.7\% and 39.8\% of their unseen samples aged 5 or younger. AgeDB spans a wider range, but retains only 95 unseen test images after exclusivity filtering.

\begin{table}[t]
    \centering
    \footnotesize
    \renewcommand{\arraystretch}{1.3}
    \setlength{\tabcolsep}{4pt}
    \begin{tabular}{lrcrrrr}
        \toprule
        & \multirow{2}{*}{\textbf{Images}} & \multirow{2}{*}{\textbf{Ages}} & \multirow{2}{*}{\makecell{\textbf{Mean} \\ \textbf{age}}} & \multicolumn{3}{c}{\textbf{Age composition}} \\
        \cmidrule(lr{0pt}){5-7}
        & & & & \textbf{0--5} & \textbf{6--12} & \textbf{13--17} \\
        \midrule
        \textbf{AFAD}     & 6,154 & 15--17 & 16.20 & 0.0\%  & 0.0\%  & 100.0\% \\
        \textbf{AgeDB}    & 95    & 3--17  & 13.76 & 5.3\%  & 21.1\% & 73.7\% \\
        \textbf{CACD2000} & 3,242 & 14--17 & 16.04 & 0.0\%  & 0.0\%  & 100.0\% \\
        \textbf{CLAP2016} & 1,127 & 1--17  & 8.30  & 39.8\% & 31.1\% & 29.1\% \\
        \textbf{UTKFace}  & 4,527 & 1--17  & 6.11  & 57.7\% & 23.8\% & 18.5\% \\
        \textbf{MORPH}    & 3,335 & 16--17 & 16.56 & 0.0\%  & 0.0\%  & 100.0\% \\
        \bottomrule
    \end{tabular}
    \caption{Age composition of the unseen test split per dataset. \textbf{Images} is the size of the unseen test split. All figures are computed from the released benchmark annotations, that is, from the images the protocol evaluates on, after split assignment and face detection.}
    \label{tab:test-split-composition}
\end{table}

This has two implications for how the benchmark should be used. The unseen MAE obtained on MORPH and the unseen MAE obtained on UTKFace do not measure the same ability: on MORPH the model extrapolates two years past the training boundary, while on UTKFace it must extrapolate into early childhood, almost three decades below the mean training age. Per-dataset scores are thus comparable only within a dataset, and our aggregate figures should be understood as an average over six distinct extrapolation problems.

The second implication concerns coverage. AFAD, CACD2000 and MORPH are the three largest datasets in the benchmark, and none of them evaluates generalization to young children. Results restricted to these datasets would give an optimistic picture of how current methods handle the age range that motivates the constraint. UTKFace and CLAP2016 are the only protocols that test that range, despite being considerably smaller.

We inherit this unevenness from the source datasets. Because the benchmark is assembled by reusing existing corpora, it requires no new data collection and no further exposure of children's images, and the unseen coverage cannot be balanced without collecting the kind of data the protocol is designed to avoid.

\section{Training Configuration}

\subsection{Hyperparameter Search}

We optimize the learning rate and weight decay for each method using the Optuna framework~\cite{akiba2019optuna} through Bayesian optimization with a \textit{Tree-Structured Parzen Estimator} (TPE), with a Hyperband pruner. All searches are conducted on UTKFace --- using the splits of \citet{paplhamcvpr2024reflect} for the supervised setting and our benchmark splits for the GZSL setting --- with the validation MAE as the optimization objective for the former and the harmonic mean of seen/unseen validation MAE for the latter. Each search runs 50 trials. The learning rate is drawn log-uniformly from $[5 \cdot 10^{-6},\, 1 \cdot 10^{-3}]$ and the weight decay log-uniformly from $[1 \cdot 10^{-6},\, 1 \cdot 10^{-4}]$. The search space, number of trials, optimizer, and batch size are kept identical across all searches, ensuring a comparable tuning budget for every method.

Table~\ref{tab:supp-hparam-tuning-results} reports the resulting learning rate and weight decay for each method in both settings, alongside the best validation score obtained.

\begin{table}[h]
    \centering
    \renewcommand{\arraystretch}{1.3}
    \small
    \setlength{\tabcolsep}{4.75pt}
    \begin{tabular}{clccc}
        \toprule
        & & \makecell{\textbf{Learning} \\ \textbf{Rate}} & \makecell{\textbf{Weight} \\ \textbf{Decay}} & \makecell{\textbf{Best} \\ \textbf{Score}} \\
        \midrule
        \multirow{9}{*}{\rotatebox{90}{\textit{\textbf{Supervised Setting}}}}
        & \textbf{Regression} & $5.1067 \cdot 10^{-5}$ & $1.3163 \cdot 10^{-6}$ & \textbf{4.44} \\
        & \textbf{DEX} & $1.6633 \cdot 10^{-5}$ & $1.2666 \cdot 10^{-5}$ & 5.21 \\
        & \textbf{SORD} & $1.6633 \cdot 10^{-5}$ & $1.2666 \cdot 10^{-5}$ & 4.85 \\
        & \textbf{DLDL} & $1.6633 \cdot 10^{-5}$ & $1.2666 \cdot 10^{-5}$ & 4.84 \\
        & \textbf{DLDL-v2} & $2.0933 \cdot 10^{-5}$ & $1.5308 \cdot 10^{-5}$ & 4.59 \\
        & \textbf{OR-CNN} & $8.0412 \cdot 10^{-4}$ & $4.7896 \cdot 10^{-6}$ & 4.62 \\
        & \textbf{Mean-Var.} & $2.5500 \cdot 10^{-4}$ & $5.7896 \cdot 10^{-6}$ & 4.68 \\
        & \textbf{CORAL} & $9.0310 \cdot 10^{-4}$ & $2.3424 \cdot 10^{-5}$ & 4.70 \\
        \rule[-6pt]{0pt}{0pt}
        & \textbf{CORN} & $1.6633 \cdot 10^{-5}$ & $1.2666 \cdot 10^{-5}$ & 4.83 \\
        \hline
        \multirow{9}{*}{\rotatebox{90}{\textit{\textbf{GZSL Setting}}}}
        \rule[0pt]{0pt}{12pt}
        & \textbf{Regression} & $6.3916 \cdot 10^{-5}$ & $6.0846 \cdot 10^{-6}$ & \textbf{7.10} \\
        & \textbf{DEX} & $5.0618 \cdot 10^{-5}$ & $2.4464 \cdot 10^{-5}$ & 7.94 \\
        & \textbf{SORD} & $2.2620 \cdot 10^{-4}$ & $7.0180 \cdot 10^{-6}$ & 7.49 \\
        & \textbf{DLDL} & $4.7387 \cdot 10^{-5}$ & $9.8295 \cdot 10^{-6}$ & 7.60 \\
        & \textbf{DLDL-v2} & $3.0806 \cdot 10^{-5}$ & $2.8714 \cdot 10^{-5}$ & 7.38 \\
        & \textbf{OR-CNN} & $1.6633 \cdot 10^{-5}$ & $1.2666 \cdot 10^{-5}$ & 7.29 \\
        & \textbf{Mean-Var.} & $2.2620 \cdot 10^{-4}$ & $7.0180 \cdot 10^{-6}$ & 7.73 \\
        & \textbf{CORAL} & $9.2045 \cdot 10^{-4}$ & $9.7784 \cdot 10^{-5}$ & 7.20 \\
        & \textbf{CORN} & $7.7920 \cdot 10^{-6}$ & $9.6323 \cdot 10^{-5}$ & 7.78 \\
        \bottomrule
    \end{tabular}
    \caption{Per-method learning rate (LR), weight decay (WD), and best validation score from hyperparameter search on UTKFace. Best score is validation MAE in the supervised setting and harmonic mean of seen/unseen validation MAE in the GZSL setting.}
    \label{tab:supp-hparam-tuning-results}
\end{table}

The considerable variation in optimal hyperparameters across methods --- spanning several orders of magnitude in both learning rate and weight decay --- confirms that a shared hyperparameter configuration would disadvantage methods with fundamentally different prediction paradigms and optimization dynamics. This is one area where the protocol of \citet{paplhamcvpr2024reflect} falls short, as it adopts the same set of hyperparameters for all methods.

\subsection{Training Setup}

All methods use a ResNet50 backbone, fully trainable, initialized from ImageNet weights, with a task-specific head operating over ages 0--101. Optimization uses AdamW. Rather than a fixed number of epochs, every run trains for a fixed 12,500 iterations, with 20 validation checks spread across training, so that datasets of very different sizes receive a comparable computational budget. Batch size varies with dataset size: 64 for AgeDB and CLAP2016, 128 for UTKFace and MORPH, and 256 for AFAD and CACD2000.

Images are center-cropped to $256 \times 256$ and normalized with ImageNet statistics. Training applies a fixed augmentation pipeline: random resized crop ($p = 0.2$, scale $0.75$--$1.0$), horizontal flip ($p = 0.2$), rotation within $\pm 10^\circ$ ($p = 0.15$), and perspective, brightness/contrast, gamma, sharpen, and pixel-dropout transforms ($p = 0.05$ each).

\subsection{Reproducibility}

All experiments, hyperparameter searches, and dataset preprocessing ran on a SLURM cluster with one NVIDIA L40S GPU per job, 24~GB of system memory and 4 CPU worker processes per run. The software stack is Python 3.12, with PyTorch and Lightning for training, Optuna for hyperparameter search, albumentations for augmentation, and the RetinaFace implementation from InsightFace for face detection. All runs set a single fixed seed (123) across the Python, NumPy, and PyTorch random number generators, and enable deterministic cuDNN kernels while disabling cuDNN benchmarking. The seed is propagated to the data loaders, the experiment runner, and the Optuna sampler.

\section{Per-Method Degradation Analysis}

We report an average degradation of 46.4\% when moving from the supervised protocol to the GZSL protocol, with the worst method reaching 52.8\%. Table~\ref{tab:method-degradation} gives the full matrix. Each entry expresses the harmonic mean of the seen and unseen test MAE under GZSL as a percentage change relative to the same method's supervised test MAE.

The aggregated errors (e.g., \textit{\textbf{All Datasets}}) are computed by averaging the MAEs across methods or datasets and taking the percentage change afterwards, not by averaging the percentages in the cells. Averaging the percentages would weight every dataset equally regardless of its error scale, which favors the datasets whose supervised MAE is smallest: MORPH and AFAD sit near 3 years, while AgeDB exceeds 6, so a one-year change means very different things on either.

\begin{table*}[t]
    \centering
    \small
    \renewcommand{\arraystretch}{1.2}
    \setlength{\tabcolsep}{6pt}
    \begin{tabular}{lrrrrrrr}
        \toprule
        & \textbf{AFAD} & \textbf{AgeDB} & \textbf{CACD2000} & \textbf{CLAP2016} & \textbf{UTKFace} & \textbf{MORPH} & \textit{\textbf{All Datasets}} \\
        \midrule
        \textbf{Regression} & +13.7 & +75.2 & +82.8 & +22.9 & +56.9 & +33.9 & \textbf{+52.1} \\
        \textbf{DEX} & +12.0 & +66.2 & +78.5 & +12.8 & +55.2 & $-$9.8 & \textbf{+40.6} \\
        \textbf{SORD} & +10.8 & +61.6 & +74.4 & +12.7 & +50.3 & $-$4.4 & \textbf{+38.9} \\
        \textbf{DLDL} & +14.2 & +65.1 & +75.6 & +20.1 & +49.2 & +5.0 & \textbf{+43.3} \\
        \textbf{DLDL-v2} & +12.9 & +68.3 & +84.1 & +21.5 & +55.2 & +5.0 & \textbf{+46.2} \\
        \textbf{OR-CNN} & +21.5 & +67.7 & +95.6 & +26.3 & +60.2 & +17.1 & \textbf{+52.8} \\
        \textbf{Mean-Var.} & +13.2 & +67.4 & +84.1 & +28.2 & +53.7 & +4.9 & \textbf{+47.4} \\
        \textbf{CORAL} & +17.8 & \textbf{+101.8} & +78.0 & +10.4 & +48.1 & +28.3 & \textbf{+52.8} \\
        \textbf{CORN} & +15.3 & +55.0 & +87.1 & +18.9 & +51.6 & +10.1 & \textbf{+43.3} \\
        \midrule
        \textit{\textbf{All Methods}} & \textit{+14.4} & \textit{+69.5} & \textit{+82.2} & \textit{+18.9} & \textit{+53.4} & \textit{+9.0} & \textbf{+46.4} \\
        \bottomrule
    \end{tabular}
    \caption{Percentage change in error when moving from the supervised protocol to the GZSL protocol, per method and dataset. Positive is worse. \textit{All Datasets} averages the six percentage changes of a method, \textit{All Methods} averages the nine percentage changes on a dataset, and the corner value is the mean of all cells.}
    \label{tab:method-degradation}
\end{table*}

Degradation varies much more across datasets, from $+9.0\%$ to $+82.2\%$, than across methods, from $+38.9\%$ to $+52.8\%$, and the ordering of the methods is fairly stable from one dataset to another. The largest single value is CORAL on AgeDB at $+101.8\%$, where the error slightly more than doubles. Together these point to the protocol as the source of the difficulty: none of the modeling choices represented among our baselines offers much protection against it.

Two entries are negative: on MORPH, DEX and SORD score slightly better under GZSL ($-9.8\%$ and $-4.4\%$) than under supervision. We do not read this as successful extrapolation. MORPH has the narrowest unseen split in the benchmark, covering only ages 16--17 (Table~\ref{tab:test-split-composition}), so a prediction near the seen-class mean already falls close to the correct value, while the supervised protocol evaluates over the full 16--77 range and is a harder regression problem in absolute terms. This is a further reason per-dataset degradation figures should not be compared with one another.

The mechanism behind this anchoring differs across the three families of methods we evaluate.

\paragraph{Classification-based methods.} Unseen ages never appear as training targets for methods such as DEX, so their logits are consistently optimized toward low probabilities. Nothing then induces the model to place probability mass on those ages at inference.

\paragraph{Ordinal and rank learning methods.} OR-CNN, CORAL and CORN decompose age estimation into binary comparisons, each predicting whether the target age exceeds a given threshold. Thresholds falling in unseen regions are trained only on seen samples, so the corresponding classifiers can settle on degenerate rules and produce systematically biased predictions.

\paragraph{Label Distribution Learning methods.} LDL methods replace one-hot targets with smooth distributions centered on the ground-truth age, propagating supervision to neighboring labels, which might be expected to help at nearby unseen ages. Our results do not bear this out. Supervision still concentrates on ages observed during training, so although the predicted distribution is smoother than a one-hot output, its expectation stays within the seen interval. Local label smoothing on its own does not support extrapolation to ages absent from training.

\section{Code and Data}

The code and data can be found at: \url{https://github.com/caiopetruccirosa/generalized-zero-shot-age-estimation}. It contains the split-construction implementation, the annotation files for the six benchmark datasets, the experiment and hyperparameter-search configurations used for the runs reported in the paper, and the aggregated result tables underlying the reported figures.

The annotation files record only image identifiers, relative paths and labels, so no facial imagery is redistributed. Regenerating the benchmark requires obtaining the six source datasets from their original providers under their respective terms; the preprocessing entry point then reproduces the partitions under the same fixed seed.

\end{document}